\documentclass[sigconf]{acmart}

\AtBeginDocument{%
}

\copyrightyear{2024}
\acmYear{2024}
\setcopyright{rightsretained}
\acmConference[KDD '24]{Proceedings of the 30th ACM SIGKDD Conference on Knowledge Discovery and Data Mining}{August 25--29, 2024}{Barcelona, Spain}
\acmBooktitle{Proceedings of the 30th ACM SIGKDD Conference on Knowledge Discovery and Data Mining (KDD '24), August 25--29, 2024, Barcelona, Spain}\acmDOI{10.1145/3637528.3671797}
\acmISBN{979-8-4007-0490-1/24/08}

\makeatletter
\gdef\@copyrightpermission{
  \begin{minipage}{0.3\columnwidth}
   \href{https://urldefense.com/v3/__https://creativecommons.org/licenses/by-nc-sa/4.0/}{\includegraphics[width=0.90\textwidth]{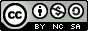}}
  \end{minipage}\hfill
  \begin{minipage}{0.7\columnwidth}
   \href{https://urldefense.com/v3/__https://creativecommons.org/licenses/by-nc-sa/4.0/}{This work is licensed under a Creative Commons Attribution-NonCommercial-ShareAlike International 4.0 License.}
  \end{minipage}
  \vspace{5pt}
}
\makeatother

\usepackage{ulem}
\usepackage{enumitem}

\usepackage{url}        

\usepackage{algorithm}
\usepackage{algpseudocode}

\usepackage{graphicx}   
\usepackage{subfigure}  
\usepackage{wrapfig, blindtext} 
\usepackage[export]{adjustbox}  

\usepackage{colortbl}
\usepackage{booktabs}       
\usepackage{multirow}       
\usepackage{multicol}       
\usepackage{array}          
\usepackage{threeparttable}

\usepackage{mathtools}
\usepackage{amsthm}
\usepackage{thmtools}

\usepackage[capitalize,noabbrev]{cleveref}

\theoremstyle{plain}
\newtheorem{theorem}{Theorem}[section]

\theoremstyle{definition}
\newtheorem{definition}[theorem]{Definition}

\newtheorem{problem}[theorem]{Problem}
\theoremstyle{remark}
\newtheorem{remark}[theorem]{Remark}

\newcommand{\beqa}{\begin{eqnarray}}
\newcommand{\eeqa}{\end{eqnarray}}
\newcommand{\beq}{\begin{equation}}
\newcommand{\eeq}{\end{equation}}
\newcommand{\ben}{\begin{enumerate}}
\newcommand{\een}{\end{enumerate}}
\newcommand{\bit}{\begin{itemize}}
\newcommand{\eit}{\end{itemize}}
\newcommand{\bi}{\begin{itemize} \item}
\newcommand{\ei}{\end{itemize}}

\newcommand{\begindef}{\begin{Definition} \rm}
\newcommand{\beginexa}{\begin{Example} \rm}
\newcommand{\beginthe}{\begin{Theorem} \rm}
\newcommand{\beginpro}{\begin{Proposition} \rm}
\newcommand{\beginlem}{\begin{Lemma} \rm}
\newcommand{\begincon}{\begin{Conjecture} \rm}
\newcommand{\begincor}{\begin{Corollary} \rm}

\newcommand{\eat}[1]{}
\def\papernumber #1 raised #2 {
\vspace{-#2}
\vbox to 0pt{\hfill\framebox{\bf Paper Number #1}}
\vspace{#2}
}

\newcommand{\hide}[1]{}

\newcommand{\adult}{\textit{Adult}\xspace}
\newcommand{\compas}{\textit{Compas}\xspace}
\newcommand{\lsa}{\textit{LSA}\xspace}
\newcommand{\meps}{\textit{MEPS}\xspace}

\usepackage{stmaryrd}

\usepackage{amsmath,amsfonts,bm,bbm}

\def\ind{{\mathbbm{1}}}

\def\eqref#1{equation~\ref{#1}}

\def\1{\mathbf{1}}

\def\vb{{\mathbf{b}}}

\def\vd{{\mathbf{d}}}

\def\vr{{\mathbf{r}}}

\def\vx{{\mathbf{x}}}

\def\mA{{\mathbf{A}}}

\def\mD{{\mathbf{D}}}

\def\mI{{\mathbf{I}}}

\def\mQ{{\mathbf{Q}}}

\def\mW{{\mathbf{W}}}

\DeclareMathAlphabet{\mathsfit}{\encodingdefault}{\sfdefault}{m}{sl}
\SetMathAlphabet{\mathsfit}{bold}{\encodingdefault}{\sfdefault}{bx}{n}

\def\gD{{\mathcal{D}}}

\def\gS{{\mathcal{S}}}

\def\gX{{\mathcal{X}}}
\def\gY{{\mathcal{Y}}}

\newcommand{\E}{\mathbb{E}}

\newcommand{\R}{\mathbb{R}}

\DeclareMathOperator*{\argmin}{arg\,min}

\usepackage{amsthm}

\definecolor{em}{gray}{0.9}
\newcommand{\cellem}{\cellcolor{em}}
\newcommand{\name}{\textsc{AIM}\xspace}
\newcommand{\namer}{\textsc{AIM}$_\texttt{REM}$\xspace}
\newcommand{\namea}{\textsc{AIM}$_\texttt{AUG}$\xspace}
\newcommand{\namedesc}{\name (Bias \uline{A}ttribution, \uline{I}nterpretation, \uline{M}itigation)\xspace}
\renewcommand{\paragraph}[1]{\vspace{0.5em}\noindent\textbf{#1}\hspace{0.5em}}
\renewcommand{\eqref}[1]{Eq.~(\ref{#1})}

\begin{document}
\title{AIM: Attributing, Interpreting, Mitigating Data Unfairness}


\author{Zhining Liu}
\email{liu326@illinois.edu}
\orcid{0000-0003-1828-2109}
\affiliation{%
  \institution{University of Illinois Urbana-Champaign}
  \city{Urbana}
  \state{IL}
  \country{USA}
}
\author{Ruizhong Qiu}
\email{rq5@illinois.edu}
\orcid{0009-0000-3253-8890}
\affiliation{%
  \institution{University of Illinois Urbana-Champaign}
  \city{Urbana}
  \state{IL}
  \country{USA}
}
\author{Zhichen Zeng}
\email{zhichenz@illinois.edu}
\orcid{0000-0002-5534-3401}
\affiliation{%
  \institution{University of Illinois Urbana-Champaign}
  \city{Urbana}
  \state{IL}
  \country{USA}
}
\author{Yada Zhu}
\email{yzhu@us.ibm.com}
\orcid{0000-0002-3338-6371}
\affiliation{%
  \institution{IBM Research}
  \city{Yorktown Heights}
  \state{NY}
  \country{USA}
}
\author{Hendrik Hamann}
\email{hendrikh@us.ibm.com}
\orcid{0000-0001-9049-1330}
\affiliation{%
  \institution{IBM Research}
  \city{Yorktown Heights}
  \state{NY}
  \country{USA}
}
\author{Hanghang Tong}
\email{htong@illinois.edu}
\orcid{0000-0003-4405-3887}
\affiliation{%
  \institution{University of Illinois Urbana-Champaign}
  \city{Urbana}
  \state{IL}
  \country{USA}
}

\renewcommand{\shortauthors}{Zhining Liu et al.}

\begin{abstract}
Data collected in the real world often encapsulates historical discrimination against disadvantaged groups and individuals.
Existing fair machine learning (FairML) research has predominantly focused on mitigating discriminative bias in the model prediction, with far less effort dedicated towards exploring how to trace biases present in the data, despite its importance for the transparency and interpretability of FairML.
To fill this gap, we investigate a novel research problem: discovering samples that reflect biases/prejudices from the training data..
Grounding on the existing fairness notions, we lay out a sample bias criterion and propose practical algorithms for measuring and countering sample bias.
The derived bias score provides intuitive \textbf{sample-level} \textbf{attribution} and \textbf{explanation} of historical bias in data.
On this basis, we further design two FairML strategies via sample-bias-informed minimal data editing. 
They can \textbf{mitigate both group and individual unfairness} at the cost of \textbf{minimal or zero predictive utility loss}.
Extensive experiments and analyses on multiple real-world datasets demonstrate the effectiveness of our methods in explaining and mitigating unfairness.
Code is available at \uline{\url{https://github.com/ZhiningLiu1998/AIM}}.
\end{abstract}




\maketitle

\section{Introduction}\label{sec:intro}

Machine learning techniques are increasingly used in high-stake scenarios such as loans, recruitment, and policing strategies.
Despite the benefits of automated decision-making, data-driven models are susceptible to biases that render their decisions potentially unfair, reflecting racism, ageism, and sexism \citep{mehrabi2021survey,caton2020survey}.
With the increasing demand for equitable and responsible use of artificial intelligence, Fair Machine Learning (FairML) has gained significant attention in both research and practice \citep{barocas2023survey,kang2021netfair}.

\begin{figure}[t]
    \centering
    \includegraphics[width=\linewidth]{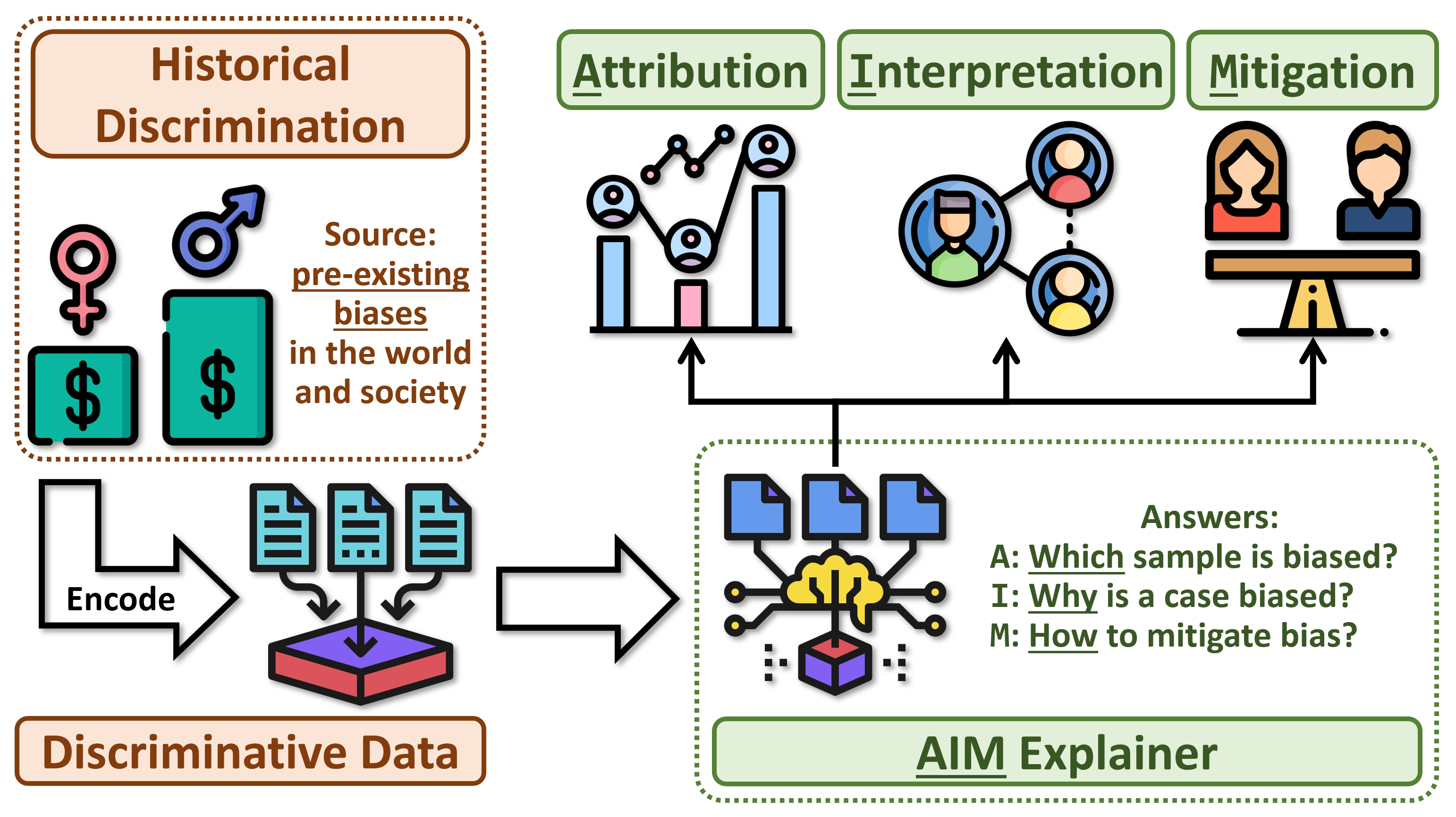}
    \vspace*{-15pt}
    \caption{
        Concept and applications of the proposed \namedesc framework.
    }
    \label{fig:intro}
    \vspace*{-10pt}
\end{figure}

Existing FairML research mainly focuses on how to devise learning algorithms to guarantee that the model has no prejudice or favoritism toward an individual or group based on their inherent or acquired characteristics \citep{mehrabi2021survey}.
In essence, the bias in machine learning models is inherited from their training data: even with perfect sampling and feature selection, data collected from the real world inevitably contains historical bias \citep{suresh2019framework,mehrabi2021survey}, which is a result of the pre-existing biases and socio-technical issues in the world \citep{kim2016data,favaretto2019bigdata}.
For example, discrimination against female/minority by loan providers recorded in databases can be inherited by the learning model for loan screening, and eventually leads to outcomes that are unfavorable for female/minority applicants \citep{fu2021loanbias}.
Similar situations arise in various domains of daily living, such as employment, housing, insurance, credit scoring, and many more~\citep{favaretto2019bigdata,fu2021loanbias}.
Identifying and understanding such biases encoded in the data is therefore crucial for achieving more transparent, interpretable, and equitable machine learning systems.

To this end, we delve into an under-explored aspect of FairML: discovering samples that reflect biases/prejudices from the training data.
Practically, this can assist human experts in understanding the bias within the data, locating and scrutinizing discriminatory samples, and developing more trustworthy FairML techniques based on these insights.
With its root in social, ethical and legal literature on fairness~\citep{suresh2019framework,fleisher2021whats,binns2020conflict,dwork2012indfairness}, we posit that a good sample bias criterion should be able to \textit{robustly capture various prejudices encoded in the data, whether targeted towards specific individuals or demographic groups.}
Grounding on this, we consider a sample exhibits bias if its comparable samples from other groups receive \textit{different} and \textit{credible} treatments.
As a practical example, we consider a female applicant being rejected for a loan as being discriminated if (1) a male applicant with similar conditions gets the loan ({\em different}), and (2) the approval is not by chance ({\em credible}).
Intuitively, it captures both individual-level and group-level fairness and prevents incidental events or noise in the real world from disturbing bias estimates.

Our criterion is grounded in the overarching principles behind popular algorithmic fairness notions such as group~\citep{dwork2012indfairness,chan2024group}, individual~\citep{dwork2012indfairness,hashimoto2018indfairness}, and counterfactual fairness~\citep{kusner2017cffairness}, but without depending on intricate causal modeling or requiring additional expert knowledge.
Existing fairness research also largely overlooks the imperfections in the data: due to various unexpected subjective and objective factors in data generation and processing, the data collected from the real world often contains unavoidable noise and errors~\citep{han2020noise,frenay2013noise}, which can significantly disturb the FairML process~\citep{xu2021robustfair,wang2020robustfair}.
To address this, we introduce sample credibility in FairML to obtain robust bias estimates, and practical algorithms are further proposed for estimating sample bias and credibility.
Our bias criterion is also self-explanatory: the bias of a sample can be naturally explained by corresponding other-group samples that receive different and credible treatments, which provides further insights for human experts to inspect the bias in the data.

Our bias and confidence estimation require a sample similarity metirc.
In principle, any reasonable similarity measure on the feature space can seamlessly integrate with our framework. 
However, due to the complexity of real-world data, practical applications typically require human experts to manually design/annotate similarities for each task, resulting in significant costs~\citep{dwork2020abstracting, mukherjee2020learnsim, fleisher2021whats}.
To this end, we propose a practical and intuitive similarity measurement that requires the minimal user input based on two key concepts: (i) creating a comparability graph to capture local similarities between input samples; and (ii) applying graph proximity measures on the comparability graph to capture global similarities reflecting the manifold structure of the input data.
This approach yields \textit{localized}, \textit{intuitive}, and \textit{interpretable} sample similarities to better support practical bias attribution and interpretation.

Finally, we explore the potential of mitigating unfairness based on bias attribution. 
We propose two strategies to mitigate unfairness with informed minimal data editing, namely unfairness removal and fairness augmentation. 
By discarding a small fraction of samples with high bias or augmenting samples with low bias, the proposed methods can \textit{mitigate group and individual unfairness} at the cost of \textit{minimal or zero predictive utility loss}.
Extensive experiments and analyses on multiple real-world FairML tasks demonstrate the effectiveness of the proposed unfairness mitigation strategies.

Figure \ref{fig:intro} shows the concept and applications of AIM.
To sum up, our \namedesc framework can assist practitioners in addressing the following problems:
\begin{itemize}
    \item \textbf{Attribution}: \textbf{Which} samples exhibit bias? 
    \item \textbf{Interpretation}: \textbf{Why} is a particular sample biased? 
    \item \textbf{Mitigation}: \textbf{How} to counter unfairness with auditable data editing and minimal utility loss?
\end{itemize}
Our contributions are 3-fold:
\begin{enumerate}
\item \textbf{Problem Formulation.} 
We formulate a novel problem of identifying samples that encode discrimination in the data, which is crucial for achieving more transparent FairML. 
\item \textbf{Algorithm Design.} 
We propose a novel framework \name. 
Armed with credibility-aware sample bias criterion and similarity based on user-defined comparable constraints, \name offers robust, practical, and self-explanatory sample-level bias attribution.
The results further support efficient unfairness mitigation with minimal data editing and utility loss.
\item \textbf{Experimental Evaluation.} We provide comprehensive experiments and analyses on real-world datasets to validate the effectiveness of \name in explaining and mitigating unfairness.
\end{enumerate}

\section{Preliminaries}\label{sec:prelim}

\paragraph{Notations.}
Real-world data usually contains both numerical (e.g., age, income) and categorical features (e.g., job type, residence).
In this paper, we consider the attribute vector $\vx$ contains numerical features with real values $\vr = [\vr^{(1)}, \vr^{(2)}, \cdots, \vr^{(n_r)}]$, and categorical features with discrete values $\vd = [\vd^{(1)}, \vd^{(2)}, \cdots, \vd^{(n_d)}]$, where $n_r$/$n_d$ denote the number of numerical/categorical features.
We use $\vr_\vx$/$\vd_\vx$ to denote the numerical/categorical part of feature vector $\vx$.
For simplicity, we assume numerical features $\vr$ contain values in $[0, 1]$ after some scaling/normalization operation.
We consider binary labels and demographic groups following the common setting in algorithm fairness \citep{mehrabi2021survey,barocas2023survey,caton2020survey}.
Without loss of generality, we consider label space $\gY:= \{0, 1\}$ and sensitive group membership space $\gS:= \{0, 1\}$, with positive value representing the favorable outcome/treatment (e.g., loan approval) and the advantaged group (e.g., gender/race with favoritism).
A dataset with $n$ samples is denoted as $\gD: \{(\vx_i, y_i, s_i) | i=0, 1, \cdots, n\}$ with the $i$-th data instance in $\gD$ as $(\vx_i, y_i, s_i)$.
We provide the problem definition as follows:
\begin{problem}[\textit{Unfairness Attribution, Interpretation, Mitigation}]\label{prb:aim}
Given a tabular dataset $\gD: \{(\vx_i, y_i, s_i) | i=0, 1, \cdots, n\}$ containing historical discrimination, we aim to solve the following problems.
\textit{Unfairness Attribution}: quantifying the historical bias carried by each instance $(\vx, y, s)$.
\textit{Unfairness Interpretation}: providing samples as justifications to explain why a specific instance $(\vx, y, s)$ is biased/unbiased.
\textit{Unfairness Mitigation}: debiasing the dataset such that the predictive model trained on the debiased $\gD$ inherits as little discrimination as possible while retaining the predictive utility.
\end{problem}
\section{Methodology}\label{sec:method}
In this section, we present our \namedesc framework.
We first introduce our sample bias criterion for unfairness \textit{attribution} and \textit{interpretation}, then discuss its rationales and connections to existing fairness notions.
In short, our criterion captures both individual-level and group-level unfairness and prevents incidental events or noise in the real world from disturbing bias estimates.
Then, we propose a novel similarity measure based on user-defined comparable constraints to support reasonable attribution and interpretation of sample bias.
It allows practical, configurable, and interpretable similarity computing in complex heterogeneous feature spaces without relying on human prior moral judgment.
Finally, we propose two practical unfairness \textit{mitigation} strategies based on the bias attribution results, namely unfairness removal (\namer) and fairness augmentation (\namea).
By removing/augmenting a small fraction of unfair/fair samples, our mitigation algorithms can alleviate both group and individual unfairness with minimal utility loss.

\subsection{Sample Bias Criterion}
We first introduce the definition of sample bias and then discuss the rationale for our design.
Aligned with the philosophy that fairness is the \textit{absence of any prejudice or favoritism towards an individual or group based on their inherent or acquired characteristics} \citep{mehrabi2021survey}, our goal is to establish a sample bias definition that can effectively characterize the prejudices encoded in data, be it directed towards specific individuals or demographic groups.
Specifically, assuming given (i) an appropriate similarity function $\sigma_\gX(\cdot, \cdot): \gX \times \gX \mapsto [0, 1]$ defined on the input feature space, and (ii) the credibility $c_i \in [0,1]$ of each data instance $(\vx_i, y_i, s_i)$, we define the criterion of sample bias as follows.

\begin{definition}[\textit{Sample Bias}]\label{def:bias}
A data sample $(\vx_i, y_i, s_i)$ is biased if its similar samples specified by $\sigma_\gX(\cdot, \cdot)$ from the other sensitive group $\gD_{s_j \neq s_i} := \{(\vx_j, y_j, s_j) | s_j \neq s_i\}$ receive different (i.e., $y_j \neq y_i$) and credible (i.e., with high credibility $c_i$) treatments.
\end{definition}
In an ideal world, the credibility of samples could be assessed by domain experts reviewing each data instance. 
However, this would incur substantial costs, making it generally impractical in real-world applications. 
When human-evaluated credibility is not available, we propose to use the following more practical definition of credibility that can be straightforwardly computed:
\begin{definition}[\textit{Sample Credibility}]\label{def:cred}
A sample $(\vx_i, y_i, s_i)$ with label $y_i$ is credible if its similar samples specified by $\sigma_\gX(\cdot, \cdot)$ from the same sensitive group $\gD_{s_j=s_i}:= \{(\vx_j, y_j, s_j) | s_j = s_i\}$ received same treatments (i.e., $y_j = y_i$).
\end{definition}
\begin{remark}[\textit{Intuition and Example}]
The intuition behind Definitions~\ref{def:bias} and \ref{def:cred} is that if an individual $a$ from group $A$ received treatment $y$ with high credibility (i.e., other same-group individuals similar to $a$ also receive the same treatment $y$.), then an individual $b$ from group $B$ with similar attributes to $a$ should also receive the same treatment $y$.
For a practical example, if a male applicant $a$ is approved for a loan, and this decision is credible (i.e., not caused by random rare events or data errors), then a female applicant $b$ with similar conditions (income, education, etc.) to $a$ should also have her loan approved.
If this is not the case, then we consider $b$ was being discriminated and thus the data sample documenting her application case exhibits historical bias.
\end{remark}

\paragraph{Connection to Existing Fairness Notions.}
Our sample bias criterion is grounded in the overarching principles behind popular algorithmic fairness notions, including group~\citep{dwork2012indfairness}, individual~\citep{dwork2012indfairness,hashimoto2018indfairness}, and counterfactual fairness~\citep{kusner2017cffairness}.
Specifically, as a widely used fairness notion, group fairness (GF) promotes equitable outcomes for different groups in terms of statistics such as positive rates.
However, GF has been criticized for lacking guarantees on the treatment of individual cases \cite{dwork2012indfairness,hashimoto2018indfairness} since it is defined on the group average.
Alternatively, individual fairness (IF) is based on the consensus that ``similar individuals should be treated similarly", but the absence of group constraints makes it challenging for IF to characterize systematic discrimination and related implicit bias in data \cite{fleisher2021whats,binns2020conflict}.
Our bias is defined on individuals but goes beyond just considering the consistency of similar samples. 
It simultaneously takes into account demographic membership to ensure fairness across different groups.
Further, counterfactual fairness (CF) is based on the intuition that “an individual should receive same decision in both the actual world and a counterfactual world where the individual belonged to a different demographic group".
Nevertheless, CF generally relies on causal models that require substantial domain knowledge and cost to obtain unobserved variables and construct the associated causal graph. 
Even with such efforts, causal models can only be built under strong assumptions~\citep{kusner2017cffairness}. Interestingly, recent research suggests that CF is largely equivalent to demographic parity \cite{rosenblatt2023cfasdp}, a basic group fairness constraint.

Our bias definition is partially inspired by CF but has been reasonably simplified, and moreover, taking into account the credibility of the samples, a point that is largely overlooked by the previous works but is crucial for robust sample bias attribution.
This makes our definition not reliant on expert knowledge and strong assumptions for constructing causal models and (ways of estimating) latent variables~\citep{kusner2017cffairness,rosenblatt2023cfasdp}, while also preventing random events/noise in the real world from disrupting bias estimation.
To sum up, compared with existing fairness notions, our sample bias definition can characterize individual-level and group-level fairness \textit{in a practical, intuitive, and robust manner}, while also allowing self-explanatory bias attribution.

\subsection{Unfairness Attribution and Interpretation}

\paragraph{Computing Self-Explanatory \name Bias Score.}
We now formally describe our \name bias attribution algorithm.
To further illustrate the practical implications of our sample bias (Definition~\ref{def:bias}) and credibility (Definition~\ref{def:cred}) criterion, we present probabilistic definitions for bias and credibility based on the underlying data distribution, and show that our criterion can be seen as utilizing a weighted local regression~\citep{cleveland1988loess,cleveland2017localreg,cleveland1979localreg} to estimate sample bias/credibility on the observed data.
To start with, we give the probabilistic definition of sample bias $b_i$ for a data instance $(\vx_i, y_i, s_i)$:
\begin{equation}\label{eq:prbbias}
    b_i:=\Pr[Y=\neg y_i\mid S=\neg s_i,X = \vx_i],
\end{equation}
i.e., the true probability that a sample with identical attributes $\vx_i$ but from another sensitive group $\neg s_i$ has the opposite label $\neg y_i$.
A larger $b_i$ signifies that the sample is more likely to have received treatment inconsistent with similar samples from different groups, thus carrying more historical bias.

However, it is evident that $b_i$ cannot be directly calculated as the underlying distribution $\Pr[Y|S,X]$ is usually unknown.
Alternatively, we can utilize weighted local regression (WLR)~\citep{cleveland1979localreg} as a non-parametric statistical method to estimate $b_i$ based on the observed data.
The general idea of WLR is to do regression on samples that are in the neighborhood of the point being estimated, thereby providing a more accurate and efficient estimation of the target based on local data patterns.
Here we employ a soft variant, where the similarity $\sigma_\gX(\cdot, \cdot)$ specifies the more local samples by assigning higher weights instead of doing hard nearest-neighbor selection~\citep{cleveland1979localreg} as in the original WLR.
For robust bias estimation, we incorporate sample credibility $\hat c_j$ as the weighting function, which will be defined later in this section.
Intuitively, if a sample $j$ has low credibility, say $\hat c_j = 0$, it will have no effect in estimating the bias.
Specifically, recall that for any event $A$, we have $\Pr[A\mid S,X]=\E[\ind[A]\mid S,X]$.
We can estimate $b_i$ by doing regression w.r.t. indicators:
\begin{equation}\label{eq:regbias}
    \hat b_i:=\argmin_{b\in\R}\sum_{j\in\gD}\ind[s_j=\neg s_i]\sigma_\gX(\vx_j, \vx_i)\hat c_j(b-\ind[y_j=\neg y_i])^2,
\end{equation}
where $\ind[s_j=\neg s_i]$ implies that we use different-group samples for regression, $\sigma_\gX(\vx_j, \vx_i)$ specifies the local samples, and $\hat c_j$ serves as the weighting function which gives high weights to more credible data points.
\eqref{eq:regbias} has closed form solution
\begin{equation}\label{eq:estbias}
    \hat b_i = \frac{
    \sum_{j\in\gD} \ind[s_j = \neg s_i] \ind[y_j = \neg y_i] \hat c_j \sigma_\gX(\vx_j, \vx_i)
    }{
    \sum_{j\in\gD} \ind[s_j = \neg s_i] \hat c_j \sigma_\gX(\vx_j, \vx_i)
    } \in [0, 1],
\end{equation}
which can be seen as an realization of our Definition~\ref{def:bias} for sample bias: we consider a sample is biased if its similar (with high $\sigma_\gX(\vx_j, \vx_i)$) samples from the other sensitive group ($\ind[s_j = \neg s_i]$) receive different ($\ind[y_j = \neg y_i]$) and credible (with high $\hat c_j$) treatments.
Similarly, the sample credibility can be defined as the true probability that sample $(\vx_i, s_i)$ should having label $y_i$:
\begin{equation}\label{eq:prbcred}
    c_i:=\Pr[Y=y_i\mid S=s_i,X=x_i].
\end{equation}
A larger $c_i$ indicates that the label of this sample is more consistent with the underlying data distribution, and therefore has higher credibility.
We can estimate $c_i$ in a similar way by solving
\begin{equation}\label{eq:regcred}
    \hat c_i:=\argmin_{c\in\R}\sum_{j\in\gD}\ind[s_j=s_i]\sigma_\gX(\vx_j, \vx_i)(c-\ind[y_j=y_i])^2.
\end{equation}
The solution of \eqref{eq:regcred} gives our credibility estimation:
\begin{equation}\label{eq:estcred}
    \hat c_i = \frac{
    \sum_{j\in\gD} \ind[s_j = s_i] \ind[y_j = y_i] \sigma_\gX(\vx_j, \vx_i)
    }{
    \sum_{j\in\gD} \ind[s_j = s_i] \sigma_\gX(\vx_j, \vx_i)
    } \in [0, 1],
\end{equation}
which is also well-aligned with our sample credibility criterion given in Definition~\ref{def:cred}: 
a sample is credible if its similar (with high $\sigma_\gX(\vx_j, \vx_i)$) samples from the same group ($\ind[s_j = s_i]$) received same ($\ind[y_j = y_i]$) treatment.

\paragraph{Interpreting the \name Bias Score.}
As mentioned earlier, and as readily observed from the bias criterion (Definition~\ref{def:bias}) and estimation (\eqref{eq:estbias}), our derived \name sample bias score is self-explanatory: the bias score $\hat b$ of each sample can be naturally explained by the corresponding samples that have contributed to $\hat b$.
Specifically, as implicated by \eqref{eq:estbias}, given the bias score $\hat b_i$ of a sample $(\vx_i, y_i, s_i)$, the bias contribution of sample $(\vx_j, y_j, s_j)$ to $\hat b_i$ is
\begin{equation}
    \hat b^\texttt{contr}_{ij} = \frac{
    \ind[s_j = \neg s_i] \ind[y_j = \neg y_i] \hat c_j \sigma_\gX(\vx_j, \vx_i)
    }{
    \sum_{j\in\gD} \ind[s_j = \neg s_i] \hat c_j \sigma_\gX(\vx_j, \vx_i)
    } \in [0, 1].
\end{equation}
Intuitively, for sample $i$ from group A, a sample $j$ from group B has large bias contribution $b^\texttt{contr}_{ij}$ if it is \textit{highly similar} to $i$ (with high $\sigma_\gX(\vx_j, \vx_i)$) and \textit{highly credible} (with large $\hat c_j$), yet received \textit{different treatment} ($\ind[y_j = \neg y_i]$).
Practitioners can discover and audit discrimination present in the data by examining the bias score and interpretation of each sample.
We conduct experiments and case studies on real-world data in Section~\ref{sec:exp-att} to validate the quality and soundness of AIM bias attribution and interpretation results.

\paragraph{Practical Similarity Computation.}
We now discuss how to determine the similarity metric $\sigma_\gX(\cdot, \cdot): \gX \times \gX \mapsto [0, 1]$ in practice. 
In principle, \textit{any similarity measure that satisfies the above definition can be seamlessly integrated with our framework}.
However, finding an appropriate similarity metric is not always easy, as real-world data can exhibit complex structure in heterogeneous feature space that contains both numerical (e.g., age, income) and categorical (e.g., residence, occupation) values.
It often requires human experts to design task-specific similarity functions based on domain knowledge, or to directly judge the similarity of sample pairs in the data, both incurring significant costs in practice~\citep{fleisher2021whats,li2023antidote}.
To address this, we present an \textit{intuitive} and \textit{practical} similarity measure that requires \textit{minimum user input} based on two key ideas: 
(i) creating a comparability graph to capture the local similarity between input samples; and 
(ii) applying a graph proximity measure on the comparability graph to capture the global similarity that reflects the manifold structure of the input data.

To start with, we first define the comparability between samples by limiting the maximum allowed disparity in numerical/categorical features.
Let $\vr_\vx$/$\vd_\vx$ represent the numerical/categorical part of feature vector $\vx$, and given user-defined numerical/categorical disparity thresholds $t_r, t_d > 0$, we define sample comparability and the comparability constraint $\Psi_{t_r, t_d}: \gX \times \gX \mapsto \{0,1\}$ as follows:
\begin{definition}[\textit{Sample Comparability}]\label{def:comp}
Two samples $\vx_1$ and $\vx_2$ have comparability under thresholds $t_r$ and $t_d$ if both holds:
(i) the disparity in any numerical feature is smaller than or equal to $t_r$, and
(ii) at most $t_d$ categorical features are different.
Formally, the above conditions can be write as a comparability constraint function:
\begin{equation}\label{eq:comp}
    \Psi_{t_r, t_d}(\vx_1, \vx_2) = 
    \begin{cases}
        1 & \text{if } \quad
        \begin{aligned}
            & \Pi_{i=1}^{n_r} \ind[|\vr_{\vx_1}^{(i)} - \vr_{\vx_2}^{(i)}| \leq t_r] = 1 \\
            & \text{ and } \Sigma_{i=1}^{n_d}\ind[\vd_{\vx_1}^{(i)}\neq \vd_{\vx_2}^{(i)}]\leq t_d;
        \end{aligned} \\
        0 & \text{otherwise.}
    \end{cases}
\end{equation}
\end{definition}
Practitioners can set $\Psi_{t_r, t_d}$ based on the application scenario and feature importance to ensure semantic similarity among comparable samples.
This realization prevents costly human evaluation for a large number of sample pairs, and avoids the complexity of finding appropriate distance functions in heterogeneous feature spaces.
The sample comparability defines a comparability graph $\mA[i, j] = \Psi_{t_r, t_d}(\vx_i,\vx_j), \forall 1 \leq i, j \leq n$ over the input data based on \textit{local similarity}.
To capture the \textit{global similarity} that reflects the manifold structure of the input data, we further utilize a graph proximity measure.
In this study, we employ random walk with restart (RWR)~\citep{pan2004rwr, tong2006rwr} due to (i) its effectiveness in capturing the global graph structure and (ii) its flexibility in adjusting the locality of the similarity.
We first remove the self-loops in $\mA$, then with symmetric normalization $\Tilde{\mW} \gets \mD^{-\frac{1}{2}} \mA \mD^{-\frac{1}{2}}$ and damping factor $p\in[0,1]$, the RWR similarity matrix can be derived by $\mQ = (1 - p) (\mI - p \Tilde{\mW})^{-1}$~\citep{pan2004rwr}, a smaller $p$ means higher restart probability and thus more locality.
We refer the readers to~\citep{pan2004rwr,tong2006rwr} and references therein for more details on the properties of RWR.
With this practical similarity measure, we can use $\mQ[i,j]$ as $\sigma_\gX(\vx_j, \vx_i)$.
We now summarize the process of \name unfairness attribution in Algorithm \ref{alg:aim}.

\begin{algorithm}[t]
    \caption{\name: \texttt{Unfairness Attribution}}\label{alg:aim}
    \begin{algorithmic}[1]
    \State  \textbf{Input:} 
    Dataset $\gD: \{(\vx_i, y_i, s_i) | i=0, 1, \cdots, n\}$,
    Comparability Constraint $\Psi: \gX \times \gX \mapsto \{0, 1\}$, Damping Factor $p \in [0, 1]$;
    \State $\mA\gets[\Psi(\vx_i,\vx_j)]_{1\le i,j\le n}$ (construct comparable graph)
    \State $\Tilde{\mW} \gets \mD^{-\frac{1}{2}} \mA \mD^{-\frac{1}{2}}$ (symmetric normalization)
    \State $\mQ \gets (1 - p) (\mI - p \Tilde{\mW})^{-1}$ (obtain similarity by RWR~\citep{tong2006rwr})
    \For{$i = 1$ to $n$}
        \State $\hat c_i \gets \frac{
        \sum_{j\in\gD} \ind[s_j = s_i] \ind[y_j = y_i] \mQ[i, j]
        }{
        \sum_{j\in\gD} \ind[s_j = s_i] \mQ[i, j]
        }$ (estimate credibility);
    \EndFor
    \For{$i = 1$ to $n$}
        \State $\hat b_i \gets \frac{
        \sum_{j\in\gD} \ind[s_j = \neg s_i] \ind[y_j = \neg y_i] \hat c_j \mQ[i, j]
        }{
        \sum_{j\in\gD} \ind[s_j = \neg s_i] \hat c_j \mQ[i, j]
        }$ (estimate bias);
    \EndFor
    \State \textbf{Return:} The sample bias vector $\hat \vb: [\hat b_1, \hat b_2, \cdots \hat b_n]$;
    \end{algorithmic}
\end{algorithm}

\subsection{\name for Unfairness Mitigation}
Our bias attribution framework can also facilitate unfairness mitigation. 
We introduce two strategies for mitigating unfairness through informed minimal data editing: unfairness removal (\namer) and fairness augmentation (\namea). 
By removing a small fraction of samples exhibiting high bias or augmenting samples with low bias, these methods can effectively \textit{mitigate both group and individual unfairness} while incurring \textit{minimal to zero loss in predictive utility}.

\paragraph{\namer: Unfairness Removal.}
The first intuitive approach to mitigating data unfairness is simply to delete samples from the dataset that exhibit high bias (i.e., carry historical discriminatory information). 
This can be achieved by simply sorting the training samples $\hat b_i$ and removing the top-K samples with the highest bias scores from the training set. 
Additionally, considering the inevitable information loss from discarding samples, to achieve fairness with minimal sample removal while also alleviating the impact of class imbalance, we adaptively select a subgroup for removal based on the class distribution. 
Specifically, we remove majority class samples to alleviate class imbalance. 
If the positive class (i.e., favorable treatment) is the majority, we select samples for removal from the privileged group (e.g., gender/race with favoritism), and vice versa.
Users can control the number of removed samples through a sample removal budget $k$.
This approach is straightforward to implement and requires little additional computational cost. 

\paragraph{\namea: Fairness Augmentation.}
Despite the simplicity and effectiveness of \namer, it may still lead to some potential information loss.
Thus we further propose an augmentation-based approach to promote fairness.
Specifically, instead of discarding unfair samples, we suggest synthesizing more fair data instances through neighborhood mixup. 
This approach can augment the pattern of fair samples, compelling the model to focus more on learning the fair patterns, and thus mitigating unfairness without deleting information from the original data.
Similarly to the above, we augment the minority class in order to alleviate class imbalance.
If the positive class (i.e., favorable treatment) is the minority, we select fair samples from the protected group (i.e., gender/race being discriminated) for augmentation, and vice versa.

Existing research indicates that simple perturbation (such as adding Gaussian noise or arbitrarily changing categorical features) may generate unrealistic samples that escape the data manifold~\citep{li2023antidote}, e.g., here we quote a good example from~\citep{li2023antidote}: ``sample with age 5 or 10 but holding a doctoral degree or getting \$80K annual income''. 
To ensure the semantic coherence of synthetic samples, we propose neighborhood-based mixup for sample synthesis.
Specifically, we first use $1-\hat b_i$ as a weight (where low-bias samples have high weights) to randomly select a fair seed sample $(\vx_s, y_s, s_s)$ for augmentation. 
Then, based on the similarity $\mQ$, we choose the most similar $n$ same-group samples and randomly select one, say $(\vx_t, y_t, s_t)$, as the mixup target.
Subsequently, we sample the mixup weight $\lambda \sim \text{Uniform}(0,1)$. 
Denoting the synthetic sample as $(\vx^*, y^*, s^*)$, it has the same group membership and label as the seed, i.e., $y^* = y_s, s^* = s_s$.
For numerical features, we simply perform linear mixup $\vr_{x^*} = \lambda \vr_{x_s} + (1-\lambda) \vr_{x_t}$.
While for each categorical feature $\vd^{(i)}$, we sample value from seed/target w.r.t. a Bernoulli distribution, i.e., $\vd^{(i)}_{x^*} = \vd^{(i)}_{x_s}$ with probability $\lambda$, and $\vd^{(i)}_{x^*} = \vd^{(i)}_{x_t}$ with probability $1-\lambda$.
It is worth noting that seed and target are similar to each other, meaning that the disparity between each numerical attributes are small and only a few categorical features are different.
Such neighborhood-based mixup ensures the semantic coherence of the synthetic fair samples.
We validate on real-world data in Section~\ref{sec:exp-mit} that both \namer and \namea can mitigate group and individual unfairness with \textit{minimal or zero predictive utility loss}.
\section{Experiments and Analysis}\label{sec:exp}

\begin{figure*}[t]
    \centering
    \includegraphics[width=\linewidth]{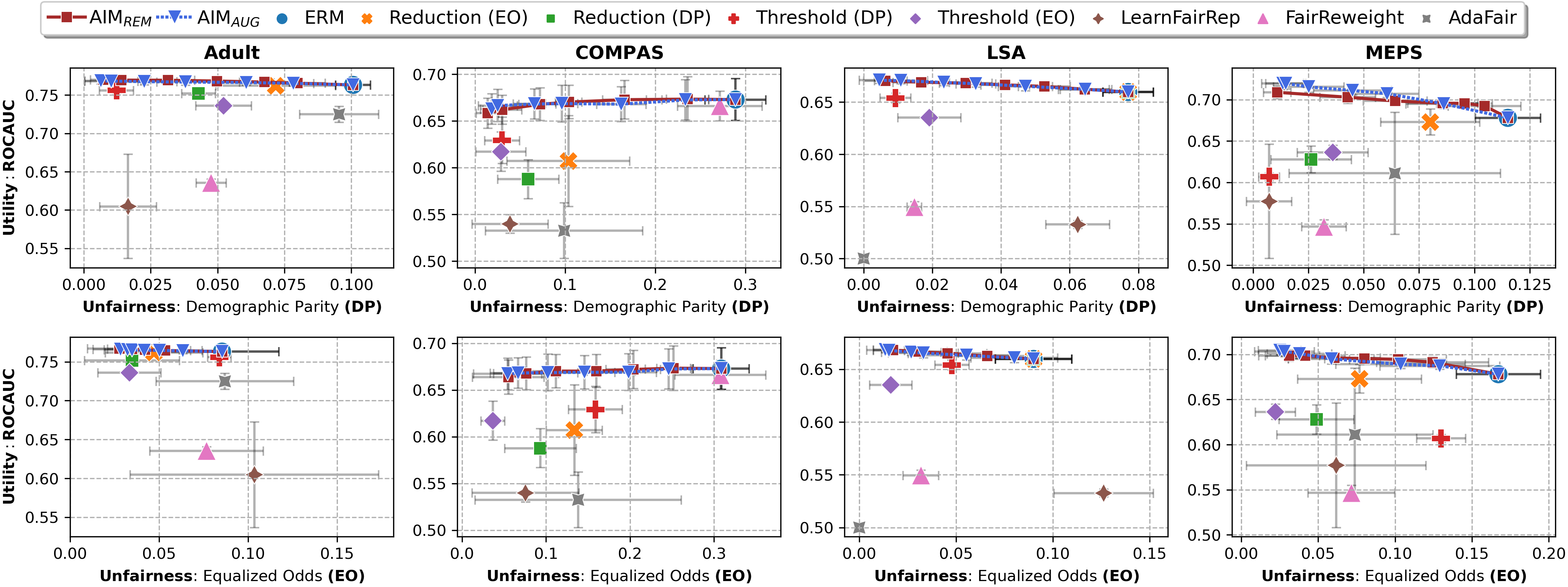}
    \vspace{-20pt}
    \caption{
        Compare \namer and \namea with group fairness baselines.
        We show the trade-off between utility (x-axis) and unfairness metrics (y-axis) on 4 real-world FairML tasks.
        Results close to the \textit{upper-left corner have better trade-offs, i.e., with low unfairness (x-axis) and high utility (y-axis).}
        Each column corresponds to a FairML task, and each row corresponds to a utility-unfairness metric pair.
        As \name's utility-unfairness trade-off can be controlled by the sample removal/augmentation budget, we show its performance with line plots. 
        We show error bars for both utility and unfairness metrics.
    }
    \label{fig:comp-group}
\end{figure*}

In this section, we conduct experiments on real-world datasets to answer the following research questions.
\begin{itemize}
    \item \textbf{RQ1 (mitigation)}: To what extent can \name alleviate various forms of discrimination against groups/individuals?
    \item \textbf{RQ2 (attribution)}: Can \name capture sample biases encoding unfair/discriminatory aspects in the data?
    \item \textbf{RQ3 (interpretation)}: How can \name provide intuitive and reasonable explanations for attributed sample biases?
\end{itemize}
We first introduce the datasets, experiment protocol, and baselines, and then present the empirical results and corresponding analysis.

\paragraph{Datasets.}
We conduct experiments on census dataset \adult~\citep{adult}, criminological dataset \compas~\citep{compas}, educational dataset \lsa~\citep{lsa} (Law School Admission), and medical dataset \meps~\citep{meps} (Medical Expenditure Panel Survey) to validate the effectiveness of the proposed \name framework in various application domains.
For each dataset, we choose one or two attributes related to ethics as sensitive attributes that exhibit significant group and individual unfairness in standard training.
More details can be found in Appendix~\ref{sec:ap-rep-data}.

\paragraph{Experiment Protocol.}
To obtain reliable results, a 5-fold cross-validation is employed and we report the average test score to eliminate randomness.
In each run, 3/1/1 folds of the data are used as the training/validation/test set (i.e., 60\%/20\%/20\% split).
We transform the categorical features into one-hot encoded features, and standardize the numerical features into the range of $[0,1]$.
We compare \name with various FairML methods proposed for group/individual fairness, considering their ability to mitigate unfairness while maintaining predictive utility.
For utility, we consider the area under the Receiver Operating Characteristic Curve (\textbf{ROC}) and Average Precision (\textbf{AP}) for unbiased utility evaluation due to class imbalance in occupation proportions in the data.
Four popular measures of group and/or individual (un)fairness are used.
For group fairness, we adopt the widely used Demographic Parity (\textbf{DP})~\citep{dwork2012indfairness} and Equalized Odds (\textbf{EO})~\citep{hardt2016eo}.
For individual fairness, we use Prediction Consistency (\textbf{PC}) following \citet{yurochkin2021sensei,yurochkin2020sensr}.
It measures the sensitivity of model to individual's group membership by testing whether $\Pr[\hat{Y}|X=x,S=s] = \Pr[\hat{Y}|X=x,S=\lnot s]$ for each test instance.
This is also known as test fairness~\citep{mehrabi2021survey} or predictive parity~\citep{chouldechova2017pi}.
We also adopt Generalized Entropy (\textbf{GE})~\citep{speicher2018ge}, a comprehensive metric that measures group and individual unfairness simultaneously with inequality indices.

\paragraph{Baselines.}
We have the following 10 FairML baselines:
(i) \textbf{Reduction}~\citep{agarwal2018reduction} reduces fair classification to a sequence of cost-sensitive classification problems, returning the classifier with the lowest empirical error subject to fair constraints.
(ii) \textbf{Threshold}~\citep{hardt2016eo} applies group-specific thresholds that optimize predictive performance while subjecting to the group fairness constraints.
(iii) \textbf{FairReweight}~\citep{kamiran2012reweight} weights the examples in each (group, label) combination differently to ensure fairness before classification.
(iv) \textbf{AdaFair}~\citep{iosifidis2019adafair} is an ensemble learning algorithm based on AdaBoost, it takes the fairness into account in each boosting round.
(v) \textbf{LearnFairRep}~\citep{zemel2013lfr} finds a latent representation which encodes the data well but obfuscates information about protected attributes. 
(vi) Sensitive Subspace Robustness (\textbf{SenSR})~\citep{yurochkin2020sensr} is an individual fairness algorithm based on Distributionally Robust Optimization (DRO). 
It finds a sensitive subspace which encodes the sensitive information most, and generates perturbations on this sensitive subspace during optimization.
(vii) Sensitive Set Invariance (\textbf{SenSeI})~\citep{yurochkin2021sensei} is also based on DRO. It involves distances penalties on both input and model predictions to construct perturbations for training individually fair model.
(viii) \textbf{FairMixup}~\citep{mroueh2021fairmixup} transforms fairness objectives into differentiable terms and optimizes them using gradient descent.
(ix) Adversarial Debiasing (\textbf{AdvFair})~\citep{adel2019advfair,zhang2018advfair} learns a classifier maximizing prediction ability while simultaneously minimizing an adversary's ability to predict sensitive attributes from predictions.
(x) Finally, \textbf{HSIC}~\cite{baharlouei2019renyi} minimizes the Hilbert-Schmidt Independence Criterion between prediction accuracy and sensitive attributes.
We consider logistic regression and neural network as base models in our experiments.
We use scikit-learn~\citep{pedregosa2011scikit} to implement logistic regression.
DRO/gradient-based FairML methods (e.g., \citep{yurochkin2020sensr,yurochkin2021sensei,mroueh2021fairmixup,adel2019advfair,baharlouei2019renyi}) that do not compatible with this pipeline will be validated with neural networks implemented with PyTorch~\citep{paszke2019pytorch}.
More implementation details can be found in Appendix~\ref{sec:ap-rep-imp}.

\vspace{-15pt}
\subsection{\name for Unfairness Mitigation}\label{sec:exp-mit}

\begin{table*}[t]
\centering
\caption{
    Compare \namer and \namea with individual fairness baselines.
    We include 3 utility and 3 (un)fairness metrics, with $\uparrow$/$\downarrow$ denoting higher/lower is better.
    For clarity, we use double/single-underline/bold to highlight the \uuline{1st}/\uline{2nd}/3rd best results.
}
\vspace{-10pt}
\label{tab:infair}
\resizebox{1.0\textwidth}{!}{%
\begin{tabular}{c|c|lc|lc|lc|lc|lc|lc}
\toprule
\multirow{3}{*}{\textbf{Task}} & \multirow{3}{*}{\textbf{Method}} & \multicolumn{6}{c|}{\multirow{2}{*}{\textbf{Utility Metrics}}} & \multicolumn{6}{c}{\textbf{Unfairness Metrics}} \\ \cline{9-14} 
 &  & \multicolumn{6}{c|}{} & \multicolumn{2}{c|}{\textbf{Unified}} & \multicolumn{2}{c|}{\textbf{Group}} & \multicolumn{2}{c}{\textbf{Individual}} \\ \cline{3-14} 
 &  & \multicolumn{1}{c}{\textbf{Acc $\uparrow$}} & \multicolumn{1}{c|}{\textbf{$\Delta$}} & \multicolumn{1}{c}{\textbf{ROC $\uparrow$}} & \multicolumn{1}{c|}{\textbf{$\Delta$}} & \multicolumn{1}{c}{\textbf{AP $\uparrow$}} & \textbf{$\Delta$} & \multicolumn{1}{c}{\textbf{GE $\downarrow$}} & \multicolumn{1}{c|}{\textbf{$\Delta$}} & \multicolumn{1}{c}{\textbf{EO $\downarrow$}} & \multicolumn{1}{c|}{\textbf{$\Delta$}} & \multicolumn{1}{c}{\textbf{PC $\uparrow$}} & \textbf{$\Delta$} \\ \hline
\multirow{9}{*}{\rotatebox{90}{gender}} & \textsc{Base} & 84.42\tiny{$\pm$0.35} & \multicolumn{1}{c|}{-} & 75.68\tiny{$\pm$1.51} & \multicolumn{1}{c|}{-} & 53.14\tiny{$\pm$1.23} & - & 8.49\tiny{$\pm$0.37} & \multicolumn{1}{c|}{-} & 11.56\tiny{$\pm$5.60} & \multicolumn{1}{c|}{\textbf{-}} & 93.90\tiny{$\pm$0.81} & - \\ \cline{2-14} 
 & \textsc{LFR} & 80.83\tiny{$\pm$3.18} & \multicolumn{1}{c|}{-4.3\%} & 66.89\tiny{$\pm$9.49} & \multicolumn{1}{c|}{-11.6\%} & 42.65\tiny{$\pm$10.07} & -19.7\% & 11.34\tiny{$\pm$2.89} & \multicolumn{1}{c|}{+33.5\%} & 13.52\tiny{$\pm$15.32} & \multicolumn{1}{c|}{+16.9\%} & 97.73\tiny{$\pm$1.59} & +4.1\% \\
 & \textsc{SenSR} & 82.65\tiny{$\pm$0.55} & \multicolumn{1}{c|}{-2.1\%} & 72.76\tiny{$\pm$1.74} & \multicolumn{1}{c|}{-3.9\%} & 48.64\tiny{$\pm$1.71} & -8.5\% & 9.59\tiny{$\pm$0.49} & \multicolumn{1}{c|}{+12.9\%} & 16.33\tiny{$\pm$3.12} & \multicolumn{1}{c|}{+41.3\%} & \uuline{\bf 99.92}\tiny{$\pm$0.05} & +6.4\% \\
 & \textsc{SenSEI} & 83.07\tiny{$\pm$0.32} & \multicolumn{1}{c|}{-1.6\%} & 72.46\tiny{$\pm$0.87} & \multicolumn{1}{c|}{-4.3\%} & 49.23\tiny{$\pm$0.83} & -7.4\% & {\bf 9.52}\tiny{$\pm$0.23} & \multicolumn{1}{c|}{+12.0\%} & 15.48\tiny{$\pm$6.42} & \multicolumn{1}{c|}{+33.9\%} & 97.61\tiny{$\pm$0.83} & +4.0\% \\
 & \textsc{AdvFair} & \uline{\bf 84.05}\tiny{$\pm$0.32} & \multicolumn{1}{c|}{-0.4\%} & {\bf 75.23}\tiny{$\pm$1.03} & \multicolumn{1}{c|}{-0.6\%} & {\bf 52.26}\tiny{$\pm$0.78} & -1.7\% & 13.91\tiny{$\pm$1.35} & \multicolumn{1}{c|}{+63.7\%} & 11.73\tiny{$\pm$7.06} & \multicolumn{1}{c|}{+1.5\%} & 92.33\tiny{$\pm$1.06} & -1.7\% \\
 & \textsc{FairMixup} & 82.66\tiny{$\pm$0.39} & \multicolumn{1}{c|}{-2.1\%} & 72.26\tiny{$\pm$1.54} & \multicolumn{1}{c|}{-4.5\%} & 48.49\tiny{$\pm$0.57} & -8.7\% & 14.37\tiny{$\pm$0.34} & \multicolumn{1}{c|}{+69.2\%} & {\bf 10.09}\tiny{$\pm$4.44} & \multicolumn{1}{c|}{-12.7\%} & 97.34\tiny{$\pm$0.50} & +3.7\% \\
 & \textsc{HSIC} & 83.45\tiny{$\pm$0.28} & \multicolumn{1}{c|}{-1.1\%} & 74.70\tiny{$\pm$0.97} & \multicolumn{1}{c|}{-1.3\%} & 50.94\tiny{$\pm$0.93} & -4.1\% & 14.09\tiny{$\pm$0.07} & \multicolumn{1}{c|}{+65.8\%} & 10.83\tiny{$\pm$3.02} & \multicolumn{1}{c|}{-6.3\%} & 97.55\tiny{$\pm$0.58} & +3.9\% \\ \cline{2-14} 
 & \cellem \namer (Ours) & \cellem {\bf 83.71}\tiny{$\pm$0.43} & {\cellem -0.8\%} & \cellem \uuline{\bf 78.25}\tiny{$\pm$1.36} & {\cellem +3.4\%} & \cellem \uline{\bf 53.28}\tiny{$\pm$1.21} & \cellem +0.3\% & \cellem \uuline{\bf 8.12}\tiny{$\pm$0.38} & {\cellem -4.4\%} & \cellem \uuline{\bf 7.30}\tiny{$\pm$2.38} & {\cellem -36.8\%} & \cellem {\bf 98.24}\tiny{$\pm$0.19} & \cellem +4.6\% \\
 & \cellem \namea (Ours) & \cellem \uuline{\bf 84.37}\tiny{$\pm$0.45} & {\cellem -0.1\%} & \cellem \uline{\bf 77.48}\tiny{$\pm$0.97} & {\cellem +2.4\%} & \cellem \uuline{\bf 53.89}\tiny{$\pm$0.99} & \cellem +1.4\% & \cellem \uline{\bf 8.15}\tiny{$\pm$0.26} & {\cellem -4.0\%} & \cellem \uline{\bf 7.64}\tiny{$\pm$1.80} & {\cellem -33.9\%} & \cellem \uline{\bf 98.69}\tiny{$\pm$0.29} & \cellem +5.1\% \\ \midrule
\multirow{9}{*}{\rotatebox{90}{race}} & \textsc{Base} & 84.38\tiny{$\pm$0.57} & \multicolumn{1}{c|}{-} & 75.70\tiny{$\pm$2.32} & \multicolumn{1}{c|}{-} & 53.08\tiny{$\pm$2.00} & - & 8.50\tiny{$\pm$0.58} & \multicolumn{1}{c|}{-} & 9.59\tiny{$\pm$4.13} & \multicolumn{1}{c|}{-} & 98.16\tiny{$\pm$0.65} & - \\ \cline{2-14} 
 & \textsc{LFR} & 80.29\tiny{$\pm$2.88} & \multicolumn{1}{c|}{-4.9\%} & 66.56\tiny{$\pm$9.39} & \multicolumn{1}{c|}{-12.1\%} & 41.64\tiny{$\pm$9.54} & -21.6\% & 11.56\tiny{$\pm$2.78} & \multicolumn{1}{c|}{+36.0\%} & 6.81\tiny{$\pm$4.32} & \multicolumn{1}{c|}{-29.0\%} & 93.20\tiny{$\pm$4.09} & -5.1\% \\
 & \textsc{SenSR} & 82.69\tiny{$\pm$0.31} & \multicolumn{1}{c|}{-2.0\%} & 71.56\tiny{$\pm$0.57} & \multicolumn{1}{c|}{-5.5\%} & 48.13\tiny{$\pm$0.66} & -9.3\% & 9.81\tiny{$\pm$0.16} & \multicolumn{1}{c|}{+15.4\%} & \uuline{\bf 6.02}\tiny{$\pm$3.92} & \multicolumn{1}{c|}{-37.2\%} & \uuline{\bf 99.91}\tiny{$\pm$0.10} & +1.8\% \\
 & \textsc{SenSEI} & 83.07\tiny{$\pm$0.31} & \multicolumn{1}{c|}{-1.6\%} & 72.67\tiny{$\pm$1.11} & \multicolumn{1}{c|}{-4.0\%} & 49.33\tiny{$\pm$0.85} & -7.1\% & {\bf 9.47}\tiny{$\pm$0.25} & \multicolumn{1}{c|}{+11.5\%} & 10.87\tiny{$\pm$2.82} & \multicolumn{1}{c|}{+13.4\%} & 98.36\tiny{$\pm$0.63} & +0.2\% \\
 & \textsc{AdvFair} & \uuline{\bf 84.61}\tiny{$\pm$0.49} & \multicolumn{1}{c|}{+0.3\%} & {\bf 77.05}\tiny{$\pm$1.30} & \multicolumn{1}{c|}{+1.8\%} & \uline{\bf 54.11}\tiny{$\pm$1.25} & +1.9\% & 14.58\tiny{$\pm$0.44} & \multicolumn{1}{c|}{+71.6\%} & 7.77\tiny{$\pm$5.30} & \multicolumn{1}{c|}{-19.0\%} & 92.99\tiny{$\pm$0.85} & -5.3\% \\
 & \textsc{FairMixup} & 82.73\tiny{$\pm$0.47} & \multicolumn{1}{c|}{-2.0\%} & 70.91\tiny{$\pm$2.08} & \multicolumn{1}{c|}{-6.3\%} & 47.94\tiny{$\pm$1.69} & -9.7\% & 14.44\tiny{$\pm$0.46} & \multicolumn{1}{c|}{+69.9\%} & 7.40\tiny{$\pm$3.22} & \multicolumn{1}{c|}{-22.8\%} & 97.12\tiny{$\pm$1.39} & -1.1\% \\
 & \textsc{HSIC} & 83.42\tiny{$\pm$0.43} & \multicolumn{1}{c|}{-1.1\%} & 75.88\tiny{$\pm$0.74} & \multicolumn{1}{c|}{+0.2\%} & 51.52\tiny{$\pm$0.96} & -2.9\% & 13.94\tiny{$\pm$0.08} & \multicolumn{1}{c|}{+64.0\%} & 7.12\tiny{$\pm$2.49} & \multicolumn{1}{c|}{-25.7\%} & 99.10\tiny{$\pm$0.40} & +1.0\% \\ \cline{2-14} 
 & \cellem \namer (Ours) & \cellem \uline{\bf 84.45}\tiny{$\pm$0.56} & {\cellem +0.1\%} & \cellem \uline{\bf 77.06}\tiny{$\pm$1.57} & {\cellem +1.8\%} & \cellem {\bf 53.81}\tiny{$\pm$1.62} & \cellem +1.4\% & \cellem \uline{\bf 8.22}\tiny{$\pm$0.44} & {\cellem -3.2\%} & \cellem {\bf 6.53}\tiny{$\pm$2.96} & {\cellem -31.9\%} & \cellem {\bf 99.21}\tiny{$\pm$0.25} & \cellem +1.1\% \\
 & \cellem \namea (Ours) & \cellem {\bf 84.38}\tiny{$\pm$0.47} & {\cellem -0.0\%} & \cellem \uuline{\bf 77.97}\tiny{$\pm$0.56} & {\cellem +3.0\%} & \cellem \uuline{\bf 54.17}\tiny{$\pm$0.86} & \cellem +2.0\% & \cellem \uuline{\bf 8.05}\tiny{$\pm$0.17} & {\cellem -5.3\%} & \cellem \uline{\bf 6.37}\tiny{$\pm$2.47} & {\cellem -33.6\%} & \cellem \uline{\bf 99.27}\tiny{$\pm$0.18} & \cellem +1.1\% \\ \bottomrule
\end{tabular}
}
\end{table*}

\paragraph{\name for group fairness (RQ1).}
We first compare \name with five FairML baselines that mitigate group fairness: Reduction, Threshold, FairReweight, AdaFair, and Learn Fair Representation.
For a comprehensive evaluation, we show the utility-fairness trade-off between ROC and group unfairness (DP, EO) on 4 real-world FairML tasks from different domains with race being the sensitive attribute.
Note that Reduction and Threshold require specifying the group unfairness constraint for optimization, thus we report their performance with DP and EO as target, respectively.
We use line plots to present the performance of \namer/\namea with different sample removal/augmentation budget which controls the trade-off of \name between utility and fairness.
The results are detailed in Figure~\ref{fig:comp-group}.
More empirical results with additional utility metrics on more FairML tasks can be found in Appendix~\ref{sec:ap-discussion}.

We summarize the key observations as follows:
\textbf{(i)} 
\textit{\name can mitigate unfairness with minimal/zero utility cost.}
Across all settings, \name achieves the optimal trade-off compared to other group fairness baselines: it either outperforms or matches the best baseline in terms of utility-fairness trade-off (close to the upper-left corner in Figure~\ref{fig:comp-group}).
\textbf{(ii)} 
Since \namer and \namea are designed to promote fairness while balancing class distribution, on datasets with significant class imbalance (e.g., \adult, \lsa, \meps with 24.8/27.9/17.2\% positive samples), they can \textit{mitigate unfairness without sacrificing predictive performance, and in some cases, may even enhance it} (e.g., ROC on the \meps dataset).
\textbf{(iii)} At the same fairness level, \namea generally exhibits better classification performance compared to \namer as it retains the original training set and promote fairness by adding augmented data.
We have similar observations in the following experiments on individual fairness.
However, we note that \namea requires a higher sample manipulation budget and computational cost than \namer.
\textbf{(iv)} 
We notice that some baselines may worsen specific fairness metrics. 
For example, LearnFairRep can help reduce DP but increase EO in the \adult and \lsa tasks.
This is due to the potential incompatibility between fairness notions, and we refer readers to~\citep{kleinberg2016tradeoff,selbst2019socialfair} for more related discussions.

\paragraph{\name for individual fairness (RQ1).}
We further verify the effectiveness of \name in promoting individual fairness (IF).
6 FairML baselines that mitigate individual unfairness are included: LearnFairRep (LFR), SenSR, SenSEI, AdvFair, FairMixup, HSIC.
Since these methods generally rely on gradient-based optimization, we use neural networks as the base ML model in this section.
Following the existing literature~\citep{yurochkin2021sensei,yurochkin2020sensr}, We test them on the widely used \adult dataset, with gender and race being the sensitive attribute, respectively.
To ensure a comprehensive evaluation, we adopt three utility metrics (ACC, ROC, AP) and three metrics for individual and/or group (un)fairness: PC for individual, EO for group, and GE for both.
For \namer and \namea, we select the sample removal/augmentation budget that maximizes PC (individual fairness) on the validation set.
The results are detailed in Table~\ref{tab:infair}. 

We summarize our findings as follows:
\textbf{(i)}
\textit{\textnormal{\namer} and \textnormal{\namea} simultaneously promote both individual and group fairness.}
Compared to the IF-targeted FairML algorithms, \name demonstrates competitive performance in mitigating individual unfairness, while concurrently achieving better group fairness.
\textbf{(ii)}
\textit{\textnormal{\name} exhibits significantly less (if any) utility losses.}
Baseline methods often lead to significant performance declines, whereas \name achieves minimal utility loss and, in some cases, even enhances certain performance metrics due to its ability to alleviate class imbalance. 
Taking \adult-gender as an example, compared to baseline methods with a relative ACC loss of 1.6-4.3\%, \name incurs only 0.1-0.8\% loss. 
This advantage becomes even more pronounced in ROC/AP, e.g., FairML baselines result in an AP loss of 7.4-19.7\%, while \name brings about a gain of 0.3-1.4\%.
This align with our earlier experiments on group fairness.
\textbf{(iii)}
\textit{FairML methods devised for individual fairness can potentially lead to group unfairness.}
For instance, on \adult-gender, while LFR/SenSR/SenSEI can promote individual PC, they also degrade group EO with an increase of 16.9/41.3/33.9\%, respectively.
GE that reflects ``how unequally the outcomes of an algorithm benefit different individuals or groups in a population''~\citep{speicher2018ge} better captures this aspect: baseline methods that improve PC actually degrade GE .
We refer the readers to \citep{binns2020conflict,fleisher2021whats} for more discussions on the trade-offs and conflicts between individual and group fairness. 

Finally, we notice that the majority of existing FairML methods are designed for a specific learning task and/or fairness metric(s), thus optimizing one metric might be at the (unintentional) expense of another metric.
In contrast, as demonstrated by the experiments above, \name avoids such intrinsic tension by focusing on the root of unfairness (i.e., historically biased samples) and thus can lead to a near-universal improvement across different metrics.

\subsection{\name for Bias Attribution and Interpretation}\label{sec:exp-att}

In this section, we conduct experiments and case studies to show the soundness of \name in unfairness attribution and interpretation.

\begin{figure}[h]
    \vspace{-5pt}
    \centering
    \includegraphics[width=\linewidth]{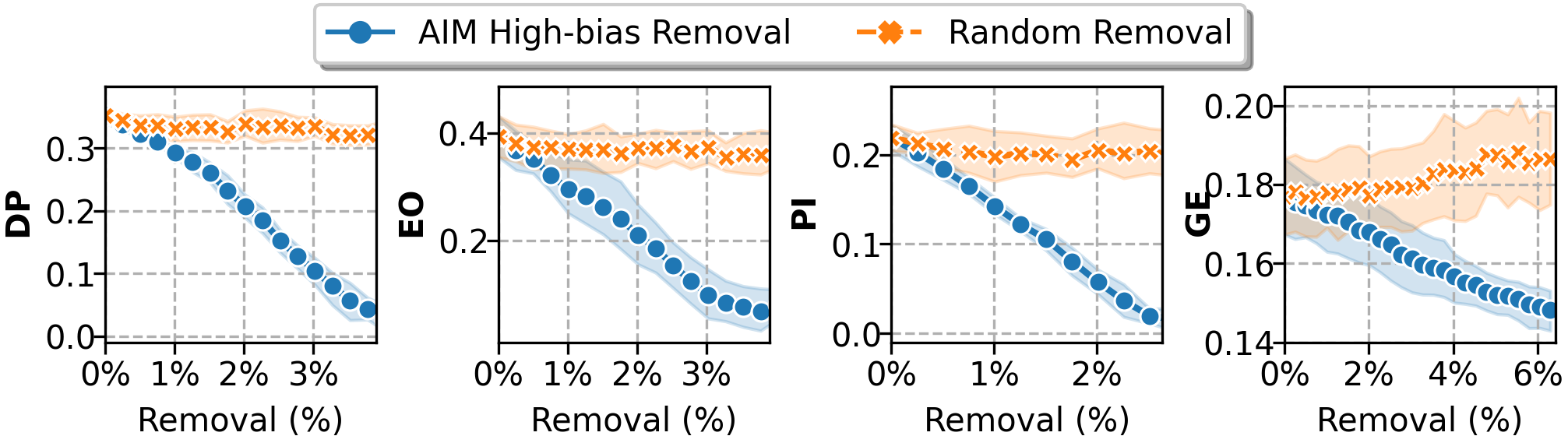}
    \vspace{-15pt}
    \caption{
        Evaluation of the \name bias attribution quality.
        Removing high-bias samples identified by \name from the data greatly reduces the discrimination in the model prediction.
    }
    \label{fig:att-rem}
    \vspace{-5pt}
\end{figure}

\paragraph{\name identifies discriminatory samples (RQ2).}
We now validate the soundness of \name unfair attribution by verifying whether the high-bias samples identified by \name encode discriminatory information from the data, and contribute to the unfairness in model predictions.
Specifically, we remove varying quantities of samples with high to low bias scores by \namer from the training dataset of \compas and observe the (un)fairness metrics of the model trained on the modified data.
We compare these results with a na\"ive method that randomly remove samples from training data.
Results are shown in Figure~\ref{fig:att-rem}.
To ensure a comprehensive evaluation, we include DP and EO for group unfairness, Predictive Inconsistency (PI, i.e., 1 - PC) for individual unfairness, and GE for both.
It can be observed that randomly removing samples does not alleviate the model's unfairness. 
At the same time, removing an equal number of high-bias samples identified by \name significantly reduces the encoded discriminatory information in the data, and effectively promotes both group and individual fairness in model predictions.
This verifies the rationality of \name unfair attribution.

\paragraph{Visual evaluation of \name bias attribution (RQ2).}
To intuitively demonstrate the bias attribution ability of \name, we further design a series of synthetic datasets with different types of bias for visual evaluation of \name's ability in capturing group- and individual-level unfairness.
In each dataset, we sample two groups from the same distribution, with group \#1 as the reference group, and introduce group/individual-level discrimination to group \#2. 
For group unfairness, we altered the class boundary for the target group to simulate group-based discrimination (different groups having different "thresholds" for positive outcomes). 
For individual unfairness, we randomly selected 10\% of samples from the target group and flipped their labels to simulate discrimination against specific individuals (similar individuals not being treated similarly). 
This approach provides ground truth labels for each sample's bias status, enabling us to visualize the quality of \name bias attribution.

\begin{figure}[h]
    \centering
    \includegraphics[width=\linewidth]{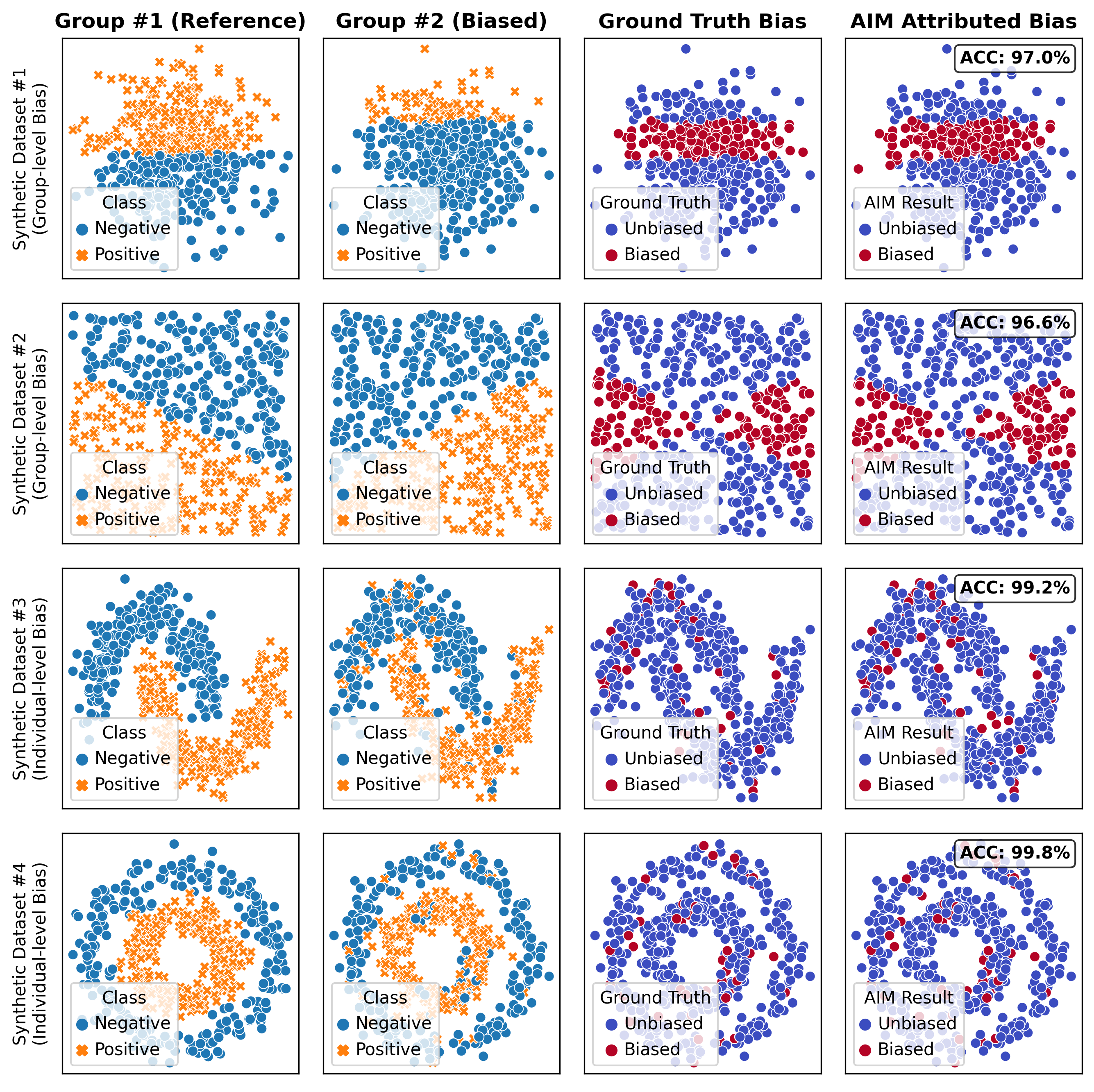}
    \vspace{-15pt}
    \caption{
        Synthetic bias detection results.
        \name (4th column) can accurately detect ground-truth biased samples (3rd columns) under both group- and individual-level unfairness.
    }
    \label{fig:syn-bias}
\end{figure}

Figure~\ref{fig:syn-bias} presents our experimental results. Each row in the figure represents a synthetic dataset, and we label the bias type of the data at the beginning of each row. For each dataset, the 1st and 2nd columns display the data and class distribution of groups \#1 and \#2, respectively. The 3rd column shows the ground truth bias distribution on the target group (\#2), where blue indicates unbiased and red indicates biased. In a similar manner, the 4th column displays the sample bias distribution detected by \name, with red indicating biased samples (with \name bias score > 0.5). 
We also annotate the accuracy of AIM-detected biased samples w.r.t. the ground truth in the 4th-column subplots. 
It can be observed that \name accurately detects group/individual-level bias in the data, with very high bias detection accuracy ranging from 97.0\% to 99.8\%.

\begin{figure}[h]
    \centering
    \includegraphics[width=\linewidth]{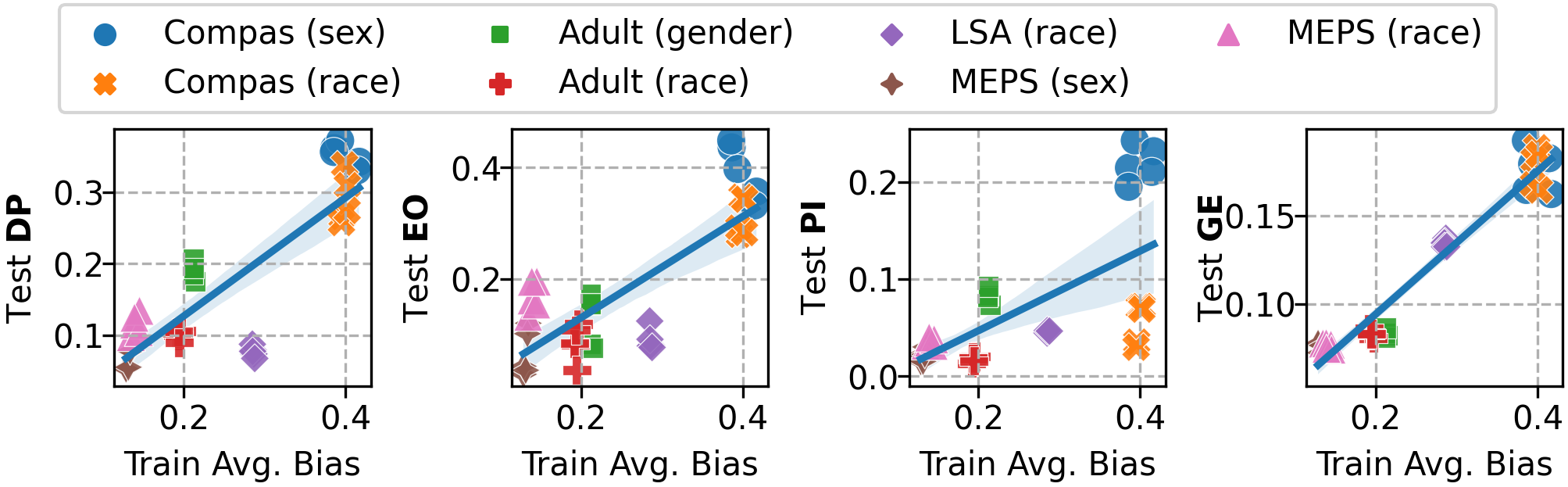}
    \vspace{-10pt}
    \caption{
        \name bias score reflect dataset unfairness level.
        Each dot denotes a combination of datasets, sensitive attributes, and train/test split.
        We report average training sample bias (x-axis) and test unfairness of the model predictions (y-axis).
    }
    \label{fig:att-dataset}
\end{figure}

\paragraph{\name reflects the level of discrimination of a dataset (RQ2).}
From a macro perspective, we further show that the outcomes of bias attribution can also offer insights into how ``discriminatory'' a dataset is.
We inherit the four metrics used in Figure~\ref{fig:att-rem}, but examine the correlation between the average \name sample bias scores on the training set and the predictive unfairness of the model on the test set, as shown in Figure~\ref{fig:att-dataset}.
It can be observed that the average \name sample bias score is a good indicator of the level of dataset discrimination, especially in terms of the comprehensive metric GE that captures both group and individual unfairness.
This further validates the soundness of \name bias attribution.

\begin{table*}[t]
\caption{Case study of \name bias attribution and interpretation on the \adult-gender task.}
\label{tab:casestudy}
\vspace{-10pt}
\begin{threeparttable}
\resizebox{\textwidth}{!}{%
\begin{tabular}{c|lllllllll|c|c|c|cc}
\toprule
\textbf{Sample} & \multicolumn{9}{c|}{\textbf{Individual Feature Attributes}} & \textbf{Sensitive} & \textbf{Label} & \textbf{Bias/Contrib} & \textbf{Explanation} & \textbf{Similarity} \\ \cline{2-13}
\textbf{Type} & \textbf{Age} & \textbf{Hours} & \textbf{CGain} & \textbf{Edu} & \textbf{Education} & \textbf{MaritalStatus} & \textbf{Occupation} & \textbf{WorkClass} & \textbf{Country} & \textbf{Sex} & \textbf{Income} & \textbf{$\hat b$ / $\hat b^\texttt{contr}$} & \textbf{Credibility $\hat c$} & \textbf{to Query} \\ \midrule
\rowcolor{em} \textbf{Query} & 32 & 45 & 5013 & 13 & Bachelors & Married-civ-spouse & Prof-specialty & Local-gov & United-States & Female & 0 & 0.9959 & - & - \\ \hline
\multirow{5}{*}{\textbf{\rotatebox{90}{Explanation}}} & 29 & 45 & 5178 & 13 & Bachelors & Married-civ-spouse & Prof-specialty & Private & United-States & Male & 1 & 0.3089 & 0.9027 & 0.3104 \\
 & 35 & 45 & 7298 & 13 & Bachelors & Married-civ-spouse & Prof-specialty & Self-emp-inc & United-States & Male & 1 & 0.2995 & 0.9916 & 0.2740 \\
 & 35 & 45 & 7298 & 13 & Bachelors & Married-civ-spouse & Prof-specialty & Private & United-States & Male & 1 & 0.2066 & 0.9822 & 0.1908 \\
 & 33 & 47 & 3103 & 13 & Bachelors & Married-civ-spouse & Prof-specialty & Private & United-States & Male & 1 & 0.0455 & 0.7136 & 0.0579 \\
 & 33 & 43 & 3103 & 13 & Bachelors & Married-civ-spouse & Prof-specialty & Private & United-States & Male & 1 & 0.0317 & 0.6546 & 0.0439 \\ \midrule
\rowcolor{em} \textbf{Query} & 34 & 40 & 5178 & 10 & Some-college & Married-civ-spouse & Prof-specialty & Private & United-States & Male & 1 & 0.7091 & - & - \\ \hline
\multirow{5}{*}{\textbf{\rotatebox{90}{Explanation}}} & 35 & 40 & 7443 & 10 & Some-college & Divorced & Prof-specialty & Private & United-States & Female & 0 & 0.4644 & 0.9875 & 0.3114 \\
 & 31 & 40 & 3908 & 10 & Some-college & Married-civ-spouse & Prof-specialty & Private & United-States & Female & 0 & 0.2130 & 0.5522 & 0.2555 \\
 & 38 & 44 & 5721 & 10 & Some-college & Divorced & Adm-clerical & Private & United-States & Female & 0 & 0.0126 & 0.8419 & 0.0099 \\
 & 32 & 40 & 2597 & 10 & Some-college & Married-civ-spouse & Exec-managerial & Private & Japan & Female & 0 & 0.0018 & 0.7789 & 0.0015 \\
 & 38 & 40 & 7443 & 10 & Some-college & Divorced & Adm-clerical & Private & United-States & Female & 0 & 0.0016 & 0.5340 & 0.0019 \\ \bottomrule
\end{tabular}
}
\begin{tablenotes}
    \scriptsize
    \item[] * Hours: working hours per week; CGain: capital gain; Edu: education-num. We show key features with differences here due to space limitation, please refer to \citep{adult} for detailed semantics of each feature.
\end{tablenotes}
\end{threeparttable}
\end{table*}

\paragraph{\name provides reasonable sample bias explanation (RQ3).}
Finally, we provide case studies on real-world data to intuitively demonstrate the validity of AIM's bias attribution and explanation. 
In Table~\ref{tab:casestudy}, we present two high-bias samples detected by AIM from different genders in the \textit{Adult-Gender} task, along with their corresponding top-5 explanations (i.e., the top 5 samples contributing most to their bias). 
It can be observed that for high-bias samples, AIM can retrieve samples from another group that are similar but have different and credible labels, and quantify their contributions to the bias ($\hat b^\texttt{contr}$) as explanations.

For example, the bias score of the first query (a female with a negative label) is largely contributed by the first male sample in line 2, who has highly similar attributes but also a different and highly credible label (with $\hat c = 0.9027$).
Similarly, for the second query sample (a male with a positive label), it is considered biased because similar females have negative labels. 
This also underscores another practical implication of the \name bias score: for a sample receiving positive/negative treatment, a high bias score suggests that it may have received an \textit{unfair advantage/disadvantage due to its group membership compared to individuals with similar attributes}.
Such sample-level explanation can assist human experts inspect and understand discrimination present in the data, and provide insights into how to design fairer decision criteria in the future.
\section{Related Works}\label{sec:related}

\paragraph{Fair Machine Learning.}
FairML advocates for ethical regulations to rectify algorithms, ensuring non-discrimination against any group or individual~\citep{caton2020survey,barocas2023survey,mehrabi2021survey}.
The concept of group fairness (GF) seeks equalized outcomes across sensitive groups concerning statistics like positive rate~\citep{hardt2016eo}.
Although intuitive, GF falls short in ensuring fairness on an individual level~\cite{dwork2012indfairness,hashimoto2018indfairness}. 
Individual fairness (IF) is thus proposed, with its main idea being that similar samples should receive similar treatment~\cite{dwork2012indfairness}.
However, due to the absence of group constraints, IF cannot capture systematic bias against groups~\citep{fleisher2021whats}.
Counterfactual fairness (CF) examines the consistency of algorithms on a single instance and its counterfactuals when sensitive attributes are altered~\citep{kusner2017cffairness}. 
However, this notion and its evaluations heavily depend on the causal structure rooted in the data generation process, thus explicit modeling is usually impractical~\citep{li2023antidote}.
AIM is grounded on the existing fairness notions and can practically capture various prejudices encoded in the data.

\paragraph{Discrimination Discovery.}
There are a few works on discrimination discovery, but their scope significantly differs from ours. 
The discrimination discovery in~\citep{ruggieri2010discover} is aimed at systems based on classification (association) rules. 
They propose an extended lift measure to assess whether a classification rule might lead to unlawful discrimination against groups protected by law, and further propose a system for  discrimination discovery in database~\citep{ruggieri2010dcube}.
Apparently, this type of discrimination discovery cannot be generalized to non-rule-based ML models.
Another line of work~\citep{zhang2017cfremove,zhang2017cfanti} models the direct/indirect discrimination as the path-specific effects on the causal network.
Like counterfactual fairness~\cite{kusner2017cffairness}, this also requires explicit modeling of the causal structure and data generation process.
Additionally, their discrimination discovery is achieved by computing a score for the entire dataset, where a high score indicates that the dataset as a whole exhibits discrimination.
This is fundamentally different from our sample-level bias attribution.
\section{Conclusion}\label{sec:con}
In this work, we investigate the problem of identifying samples carrying historical biases in training data. 
Building on existing fairness notions, we establish a criterion and propose practical algorithms for measuring and countering sample bias.
We propose a practical framework \name, which supports (i) sample-level bias attribution, (ii) intuitive explanation of the bias of each instance, and (iii) effective unfairness mitigation with minimal or zero predictive utility loss.
Extensive experiments and analysis on various real-world datasets demonstrate the efficacy of our approach in attributing, interpreting, and mitigating unfairness.

\begin{acks}
This work is supported by NSF (1939725), 
AFOSR (FA9550-24-1-0002), 
the C3.ai Digital Transformation Institute, 
MIT-IBM Watson AI Lab, 
and IBM-Illinois Discovery Accelerator Institute. 
The content of the information in this document does not necessarily reflect the position or the policy of the Government, and no official endorsement should be inferred.  The U.S. Government is authorized to reproduce and distribute reprints for Government purposes notwithstanding any copyright notation here on.
\end{acks}

\normalem

\bibliographystyle{ACM-Reference-Format}
\bibliography{ref}


\begin{thebibliography}{66}


\ifx \showCODEN    \undefined \def \showCODEN     #1{\unskip}     \fi
\ifx \showDOI      \undefined \def \showDOI       #1{#1}\fi
\ifx \showISBNx    \undefined \def \showISBNx     #1{\unskip}     \fi
\ifx \showISBNxiii \undefined \def \showISBNxiii  #1{\unskip}     \fi
\ifx \showISSN     \undefined \def \showISSN      #1{\unskip}     \fi
\ifx \showLCCN     \undefined \def \showLCCN      #1{\unskip}     \fi
\ifx \shownote     \undefined \def \shownote      #1{#1}          \fi
\ifx \showarticletitle \undefined \def \showarticletitle #1{#1}   \fi
\ifx \showURL      \undefined \def \showURL       {\relax}        \fi
\providecommand\bibfield[2]{#2}
\providecommand\bibinfo[2]{#2}
\providecommand\natexlab[1]{#1}
\providecommand\showeprint[2][]{arXiv:#2}

\bibitem[Adel et~al\mbox{.}(2019)]%
        {adel2019advfair}
\bibfield{author}{\bibinfo{person}{Tameem Adel}, \bibinfo{person}{Isabel Valera}, \bibinfo{person}{Zoubin Ghahramani}, {and} \bibinfo{person}{Adrian Weller}.} \bibinfo{year}{2019}\natexlab{}.
\newblock \showarticletitle{One-network adversarial fairness}. In \bibinfo{booktitle}{\emph{Proceedings of the AAAI Conference on Artificial Intelligence}}, Vol.~\bibinfo{volume}{33}. \bibinfo{pages}{2412--2420}.
\newblock


\bibitem[Agarwal et~al\mbox{.}(2018)]%
        {agarwal2018reduction}
\bibfield{author}{\bibinfo{person}{Alekh Agarwal}, \bibinfo{person}{Alina Beygelzimer}, \bibinfo{person}{Miroslav Dud{\'\i}k}, \bibinfo{person}{John Langford}, {and} \bibinfo{person}{Hanna Wallach}.} \bibinfo{year}{2018}\natexlab{}.
\newblock \showarticletitle{A reductions approach to fair classification}. In \bibinfo{booktitle}{\emph{International conference on machine learning}}. PMLR, \bibinfo{pages}{60--69}.
\newblock


\bibitem[Agarwal(2023)]%
        {infairness}
\bibfield{author}{\bibinfo{person}{Mayank Agarwal}.} \bibinfo{year}{2023}\natexlab{}.
\newblock \bibinfo{title}{Individual Fairness and inFairness}.
\newblock \bibinfo{howpublished}{\url{https://github.com/IBM/inFairness}}.
\newblock


\bibitem[Angwin et~al\mbox{.}(2016)]%
        {compas}
\bibfield{author}{\bibinfo{person}{Julia Angwin}, \bibinfo{person}{Jeff Larson}, \bibinfo{person}{Surya Mattu}, {and} \bibinfo{person}{Lauren Kirchner}.} \bibinfo{year}{2016}\natexlab{}.
\newblock \showarticletitle{Machine bias}. In \bibinfo{booktitle}{\emph{Ethics of Data and Analytics}}.
\newblock


\bibitem[Baharlouei et~al\mbox{.}(2019)]%
        {baharlouei2019renyi}
\bibfield{author}{\bibinfo{person}{Sina Baharlouei}, \bibinfo{person}{Maher Nouiehed}, \bibinfo{person}{Ahmad Beirami}, {and} \bibinfo{person}{Meisam Razaviyayn}.} \bibinfo{year}{2019}\natexlab{}.
\newblock \showarticletitle{R$\backslash$'enyi Fair Inference}.
\newblock \bibinfo{journal}{\emph{arXiv preprint arXiv:1906.12005}} (\bibinfo{year}{2019}).
\newblock


\bibitem[Barocas et~al\mbox{.}(2023)]%
        {barocas2023survey}
\bibfield{author}{\bibinfo{person}{Solon Barocas}, \bibinfo{person}{Moritz Hardt}, {and} \bibinfo{person}{Arvind Narayanan}.} \bibinfo{year}{2023}\natexlab{}.
\newblock \bibinfo{booktitle}{\emph{Fairness and machine learning: Limitations and opportunities}}.
\newblock \bibinfo{publisher}{MIT Press}.
\newblock


\bibitem[Bellamy et~al\mbox{.}(2019)]%
        {bellamy2019aif}
\bibfield{author}{\bibinfo{person}{Rachel~KE Bellamy}, \bibinfo{person}{Kuntal Dey}, \bibinfo{person}{Michael Hind}, \bibinfo{person}{Samuel~C Hoffman}, \bibinfo{person}{Stephanie Houde}, \bibinfo{person}{Kalapriya Kannan}, \bibinfo{person}{Pranay Lohia}, \bibinfo{person}{Jacquelyn Martino}, \bibinfo{person}{Sameep Mehta}, \bibinfo{person}{Aleksandra Mojsilovi{\'c}}, {et~al\mbox{.}}} \bibinfo{year}{2019}\natexlab{}.
\newblock \showarticletitle{AI Fairness 360: An extensible toolkit for detecting and mitigating algorithmic bias}.
\newblock \bibinfo{journal}{\emph{IBM Journal of Research and Development}} \bibinfo{volume}{63}, \bibinfo{number}{4/5} (\bibinfo{year}{2019}), \bibinfo{pages}{4--1}.
\newblock


\bibitem[Binns(2020)]%
        {binns2020conflict}
\bibfield{author}{\bibinfo{person}{Reuben Binns}.} \bibinfo{year}{2020}\natexlab{}.
\newblock \showarticletitle{On the apparent conflict between individual and group fairness}. In \bibinfo{booktitle}{\emph{Proceedings of the 2020 conference on fairness, accountability, and transparency}}. \bibinfo{pages}{514--524}.
\newblock


\bibitem[Bird et~al\mbox{.}(2020)]%
        {bird2020fairlearn}
\bibfield{author}{\bibinfo{person}{Sarah Bird}, \bibinfo{person}{Miro Dud{\'\i}k}, \bibinfo{person}{Richard Edgar}, \bibinfo{person}{Brandon Horn}, \bibinfo{person}{Roman Lutz}, \bibinfo{person}{Vanessa Milan}, \bibinfo{person}{Mehrnoosh Sameki}, \bibinfo{person}{Hanna Wallach}, {and} \bibinfo{person}{Kathleen Walker}.} \bibinfo{year}{2020}\natexlab{}.
\newblock \showarticletitle{Fairlearn: A toolkit for assessing and improving fairness in AI}.
\newblock \bibinfo{journal}{\emph{Microsoft, Tech. Rep. MSR-TR-2020-32}} (\bibinfo{year}{2020}).
\newblock


\bibitem[Caton and Haas(2020)]%
        {caton2020survey}
\bibfield{author}{\bibinfo{person}{Simon Caton} {and} \bibinfo{person}{Christian Haas}.} \bibinfo{year}{2020}\natexlab{}.
\newblock \showarticletitle{Fairness in machine learning: A survey}.
\newblock \bibinfo{journal}{\emph{Comput. Surveys}} (\bibinfo{year}{2020}).
\newblock


\bibitem[Chan et~al\mbox{.}(2024)]%
        {chan2024group}
\bibfield{author}{\bibinfo{person}{Eunice Chan}, \bibinfo{person}{Zhining Liu}, \bibinfo{person}{Ruizhong Qiu}, \bibinfo{person}{Yuheng Zhang}, \bibinfo{person}{Ross Maciejewski}, {and} \bibinfo{person}{Hanghang Tong}.} \bibinfo{year}{2024}\natexlab{}.
\newblock \showarticletitle{Group Fairness via Group Consensus}. In \bibinfo{booktitle}{\emph{The 2024 ACM Conference on Fairness, Accountability, and Transparency}}. \bibinfo{pages}{1788--1808}.
\newblock


\bibitem[Chouldechova(2017)]%
        {chouldechova2017pi}
\bibfield{author}{\bibinfo{person}{Alexandra Chouldechova}.} \bibinfo{year}{2017}\natexlab{}.
\newblock \showarticletitle{Fair prediction with disparate impact: A study of bias in recidivism prediction instruments}.
\newblock \bibinfo{journal}{\emph{Big data}} \bibinfo{volume}{5}, \bibinfo{number}{2} (\bibinfo{year}{2017}), \bibinfo{pages}{153--163}.
\newblock


\bibitem[Cleveland(1979)]%
        {cleveland1979localreg}
\bibfield{author}{\bibinfo{person}{William~S Cleveland}.} \bibinfo{year}{1979}\natexlab{}.
\newblock \showarticletitle{Robust locally weighted regression and smoothing scatterplots}.
\newblock \bibinfo{journal}{\emph{Journal of the American statistical association}} \bibinfo{volume}{74}, \bibinfo{number}{368} (\bibinfo{year}{1979}), \bibinfo{pages}{829--836}.
\newblock


\bibitem[Cleveland and Devlin(1988)]%
        {cleveland1988loess}
\bibfield{author}{\bibinfo{person}{William~S Cleveland} {and} \bibinfo{person}{Susan~J Devlin}.} \bibinfo{year}{1988}\natexlab{}.
\newblock \showarticletitle{Locally weighted regression: an approach to regression analysis by local fitting}.
\newblock \bibinfo{journal}{\emph{Journal of the American statistical association}} \bibinfo{volume}{83}, \bibinfo{number}{403} (\bibinfo{year}{1988}), \bibinfo{pages}{596--610}.
\newblock


\bibitem[Cleveland et~al\mbox{.}(2017)]%
        {cleveland2017localreg}
\bibfield{author}{\bibinfo{person}{William~S Cleveland}, \bibinfo{person}{Eric Grosse}, {and} \bibinfo{person}{William~M Shyu}.} \bibinfo{year}{2017}\natexlab{}.
\newblock \showarticletitle{Local regression models}.
\newblock In \bibinfo{booktitle}{\emph{Statistical models in S}}. \bibinfo{publisher}{Routledge}, \bibinfo{pages}{309--376}.
\newblock


\bibitem[Cohen et~al\mbox{.}(2009)]%
        {meps}
\bibfield{author}{\bibinfo{person}{Joel~W Cohen}, \bibinfo{person}{Steven~B Cohen}, {and} \bibinfo{person}{Jessica~S Banthin}.} \bibinfo{year}{2009}\natexlab{}.
\newblock \showarticletitle{The medical expenditure panel survey: a national information resource to support healthcare cost research and inform policy and practice}.
\newblock \bibinfo{journal}{\emph{Medical care}} (\bibinfo{year}{2009}), \bibinfo{pages}{S44--S50}.
\newblock


\bibitem[Dwork et~al\mbox{.}(2012)]%
        {dwork2012indfairness}
\bibfield{author}{\bibinfo{person}{Cynthia Dwork}, \bibinfo{person}{Moritz Hardt}, \bibinfo{person}{Toniann Pitassi}, \bibinfo{person}{Omer Reingold}, {and} \bibinfo{person}{Richard Zemel}.} \bibinfo{year}{2012}\natexlab{}.
\newblock \showarticletitle{Fairness through awareness}. In \bibinfo{booktitle}{\emph{Proceedings of the 3rd innovations in theoretical computer science conference}}. \bibinfo{pages}{214--226}.
\newblock


\bibitem[Dwork et~al\mbox{.}(2020)]%
        {dwork2020abstracting}
\bibfield{author}{\bibinfo{person}{Cynthia Dwork}, \bibinfo{person}{Christina Ilvento}, \bibinfo{person}{Guy~N Rothblum}, {and} \bibinfo{person}{Pragya Sur}.} \bibinfo{year}{2020}\natexlab{}.
\newblock \showarticletitle{Abstracting fairness: Oracles, metrics, and interpretability}.
\newblock \bibinfo{journal}{\emph{arXiv preprint arXiv:2004.01840}} (\bibinfo{year}{2020}).
\newblock


\bibitem[Favaretto et~al\mbox{.}(2019)]%
        {favaretto2019bigdata}
\bibfield{author}{\bibinfo{person}{Maddalena Favaretto}, \bibinfo{person}{Eva De~Clercq}, {and} \bibinfo{person}{Bernice~Simone Elger}.} \bibinfo{year}{2019}\natexlab{}.
\newblock \showarticletitle{Big Data and discrimination: perils, promises and solutions. A systematic review}.
\newblock \bibinfo{journal}{\emph{Journal of Big Data}} \bibinfo{volume}{6}, \bibinfo{number}{1} (\bibinfo{year}{2019}), \bibinfo{pages}{1--27}.
\newblock


\bibitem[Fleisher(2021)]%
        {fleisher2021whats}
\bibfield{author}{\bibinfo{person}{Will Fleisher}.} \bibinfo{year}{2021}\natexlab{}.
\newblock \showarticletitle{What's fair about individual fairness?}. In \bibinfo{booktitle}{\emph{Proceedings of the 2021 AAAI/ACM Conference on AI, Ethics, and Society}}. \bibinfo{pages}{480--490}.
\newblock


\bibitem[Fr{\'e}nay and Verleysen(2013)]%
        {frenay2013noise}
\bibfield{author}{\bibinfo{person}{Beno{\^\i}t Fr{\'e}nay} {and} \bibinfo{person}{Michel Verleysen}.} \bibinfo{year}{2013}\natexlab{}.
\newblock \showarticletitle{Classification in the presence of label noise: a survey}.
\newblock \bibinfo{journal}{\emph{IEEE transactions on neural networks and learning systems}} \bibinfo{volume}{25}, \bibinfo{number}{5} (\bibinfo{year}{2013}), \bibinfo{pages}{845--869}.
\newblock


\bibitem[Fu et~al\mbox{.}(2022)]%
        {DBLP:conf/kdd/FuFMTH22}
\bibfield{author}{\bibinfo{person}{Dongqi Fu}, \bibinfo{person}{Liri Fang}, \bibinfo{person}{Ross Maciejewski}, \bibinfo{person}{Vetle~I. Torvik}, {and} \bibinfo{person}{Jingrui He}.} \bibinfo{year}{2022}\natexlab{}.
\newblock \showarticletitle{Meta-Learned Metrics over Multi-Evolution Temporal Graphs}. In \bibinfo{booktitle}{\emph{KDD}}.
\newblock


\bibitem[Fu and He(2021)]%
        {DBLP:conf/sigir/FuH21}
\bibfield{author}{\bibinfo{person}{Dongqi Fu} {and} \bibinfo{person}{Jingrui He}.} \bibinfo{year}{2021}\natexlab{}.
\newblock \showarticletitle{{SDG:} {A} Simplified and Dynamic Graph Neural Network}. In \bibinfo{booktitle}{\emph{SIGIR}}.
\newblock


\bibitem[Fu et~al\mbox{.}(2020a)]%
        {DBLP:conf/cikm/FuXLTH20}
\bibfield{author}{\bibinfo{person}{Dongqi Fu}, \bibinfo{person}{Zhe Xu}, \bibinfo{person}{Bo Li}, \bibinfo{person}{Hanghang Tong}, {and} \bibinfo{person}{Jingrui He}.} \bibinfo{year}{2020}\natexlab{a}.
\newblock \showarticletitle{A View-Adversarial Framework for Multi-View Network Embedding}. In \bibinfo{booktitle}{\emph{CIKM}}.
\newblock


\bibitem[Fu et~al\mbox{.}(2020b)]%
        {DBLP:conf/kdd/FuZH20}
\bibfield{author}{\bibinfo{person}{Dongqi Fu}, \bibinfo{person}{Dawei Zhou}, {and} \bibinfo{person}{Jingrui He}.} \bibinfo{year}{2020}\natexlab{b}.
\newblock \showarticletitle{Local Motif Clustering on Time-Evolving Graphs}. In \bibinfo{booktitle}{\emph{KDD}}.
\newblock


\bibitem[Fu et~al\mbox{.}(2023)]%
        {DBLP:conf/www/Fu0MCBH23}
\bibfield{author}{\bibinfo{person}{Dongqi Fu}, \bibinfo{person}{Dawei Zhou}, \bibinfo{person}{Ross Maciejewski}, \bibinfo{person}{Arie Croitoru}, \bibinfo{person}{Marcus Boyd}, {and} \bibinfo{person}{Jingrui He}.} \bibinfo{year}{2023}\natexlab{}.
\newblock \showarticletitle{Fairness-Aware Clique-Preserving Spectral Clustering of Temporal Graphs}. In \bibinfo{booktitle}{\emph{The {ACM} Web Conference 2023, {WWW}}}.
\newblock


\bibitem[Fu et~al\mbox{.}(2021)]%
        {fu2021loanbias}
\bibfield{author}{\bibinfo{person}{Runshan Fu}, \bibinfo{person}{Yan Huang}, {and} \bibinfo{person}{Param~Vir Singh}.} \bibinfo{year}{2021}\natexlab{}.
\newblock \showarticletitle{Crowds, lending, machine, and bias}.
\newblock \bibinfo{journal}{\emph{Information Systems Research}} \bibinfo{volume}{32}, \bibinfo{number}{1} (\bibinfo{year}{2021}), \bibinfo{pages}{72--92}.
\newblock


\bibitem[Han et~al\mbox{.}(2020)]%
        {han2020noise}
\bibfield{author}{\bibinfo{person}{Bo Han}, \bibinfo{person}{Quanming Yao}, \bibinfo{person}{Tongliang Liu}, \bibinfo{person}{Gang Niu}, \bibinfo{person}{Ivor~W Tsang}, \bibinfo{person}{James~T Kwok}, {and} \bibinfo{person}{Masashi Sugiyama}.} \bibinfo{year}{2020}\natexlab{}.
\newblock \showarticletitle{A survey of label-noise representation learning: Past, present and future}.
\newblock \bibinfo{journal}{\emph{arXiv preprint arXiv:2011.04406}} (\bibinfo{year}{2020}).
\newblock


\bibitem[Hardt et~al\mbox{.}(2016)]%
        {hardt2016eo}
\bibfield{author}{\bibinfo{person}{Moritz Hardt}, \bibinfo{person}{Eric Price}, {and} \bibinfo{person}{Nati Srebro}.} \bibinfo{year}{2016}\natexlab{}.
\newblock \showarticletitle{Equality of opportunity in supervised learning}.
\newblock \bibinfo{journal}{\emph{Advances in neural information processing systems}}  \bibinfo{volume}{29} (\bibinfo{year}{2016}).
\newblock


\bibitem[Hashimoto et~al\mbox{.}(2018)]%
        {hashimoto2018indfairness}
\bibfield{author}{\bibinfo{person}{Tatsunori Hashimoto}, \bibinfo{person}{Megha Srivastava}, \bibinfo{person}{Hongseok Namkoong}, {and} \bibinfo{person}{Percy Liang}.} \bibinfo{year}{2018}\natexlab{}.
\newblock \showarticletitle{Fairness without demographics in repeated loss minimization}. In \bibinfo{booktitle}{\emph{International Conference on Machine Learning}}. PMLR, \bibinfo{pages}{1929--1938}.
\newblock


\bibitem[Iosifidis and Ntoutsi(2019)]%
        {iosifidis2019adafair}
\bibfield{author}{\bibinfo{person}{Vasileios Iosifidis} {and} \bibinfo{person}{Eirini Ntoutsi}.} \bibinfo{year}{2019}\natexlab{}.
\newblock \showarticletitle{Adafair: Cumulative fairness adaptive boosting}. In \bibinfo{booktitle}{\emph{Proceedings of the 28th ACM international conference on information and knowledge management}}. \bibinfo{pages}{781--790}.
\newblock


\bibitem[Kamiran and Calders(2012)]%
        {kamiran2012reweight}
\bibfield{author}{\bibinfo{person}{Faisal Kamiran} {and} \bibinfo{person}{Toon Calders}.} \bibinfo{year}{2012}\natexlab{}.
\newblock \showarticletitle{Data preprocessing techniques for classification without discrimination}.
\newblock \bibinfo{journal}{\emph{Knowledge and information systems}} \bibinfo{volume}{33}, \bibinfo{number}{1} (\bibinfo{year}{2012}), \bibinfo{pages}{1--33}.
\newblock


\bibitem[Kang and Tong(2021)]%
        {kang2021netfair}
\bibfield{author}{\bibinfo{person}{Jian Kang} {and} \bibinfo{person}{Hanghang Tong}.} \bibinfo{year}{2021}\natexlab{}.
\newblock \showarticletitle{Fair graph mining}. In \bibinfo{booktitle}{\emph{Proceedings of the 30th ACM International Conference on Information \& Knowledge Management}}. \bibinfo{pages}{4849--4852}.
\newblock


\bibitem[Kim(2016)]%
        {kim2016data}
\bibfield{author}{\bibinfo{person}{Pauline~T Kim}.} \bibinfo{year}{2016}\natexlab{}.
\newblock \showarticletitle{Data-driven discrimination at work}.
\newblock \bibinfo{journal}{\emph{Wm. \& Mary L. Rev.}}  \bibinfo{volume}{58} (\bibinfo{year}{2016}), \bibinfo{pages}{857}.
\newblock


\bibitem[Kleinberg et~al\mbox{.}(2016)]%
        {kleinberg2016tradeoff}
\bibfield{author}{\bibinfo{person}{Jon Kleinberg}, \bibinfo{person}{Sendhil Mullainathan}, {and} \bibinfo{person}{Manish Raghavan}.} \bibinfo{year}{2016}\natexlab{}.
\newblock \showarticletitle{Inherent trade-offs in the fair determination of risk scores}.
\newblock \bibinfo{journal}{\emph{arXiv preprint arXiv:1609.05807}} (\bibinfo{year}{2016}).
\newblock


\bibitem[Kohavi and Becker(1996)]%
        {adult}
\bibfield{author}{\bibinfo{person}{Ronny Kohavi} {and} \bibinfo{person}{Barry Becker}.} \bibinfo{year}{1996}\natexlab{}.
\newblock \bibinfo{title}{Adult Data Set}.
\newblock
\newblock


\bibitem[Kusner et~al\mbox{.}(2017)]%
        {kusner2017cffairness}
\bibfield{author}{\bibinfo{person}{Matt~J Kusner}, \bibinfo{person}{Joshua Loftus}, \bibinfo{person}{Chris Russell}, {and} \bibinfo{person}{Ricardo Silva}.} \bibinfo{year}{2017}\natexlab{}.
\newblock \showarticletitle{Counterfactual fairness}.
\newblock \bibinfo{journal}{\emph{Advances in neural information processing systems}}  \bibinfo{volume}{30} (\bibinfo{year}{2017}).
\newblock


\bibitem[Li et~al\mbox{.}(2023)]%
        {li2023antidote}
\bibfield{author}{\bibinfo{person}{Peizhao Li}, \bibinfo{person}{Ethan Xia}, {and} \bibinfo{person}{Hongfu Liu}.} \bibinfo{year}{2023}\natexlab{}.
\newblock \showarticletitle{Learning antidote data to individual unfairness}. In \bibinfo{booktitle}{\emph{International Conference on Machine Learning}}. PMLR, \bibinfo{pages}{20168--20181}.
\newblock


\bibitem[Liu et~al\mbox{.}(2020a)]%
        {liu2020self}
\bibfield{author}{\bibinfo{person}{Zhining Liu}, \bibinfo{person}{Wei Cao}, \bibinfo{person}{Zhifeng Gao}, \bibinfo{person}{Jiang Bian}, \bibinfo{person}{Hechang Chen}, \bibinfo{person}{Yi Chang}, {and} \bibinfo{person}{Tie-Yan Liu}.} \bibinfo{year}{2020}\natexlab{a}.
\newblock \showarticletitle{Self-paced ensemble for highly imbalanced massive data classification}. In \bibinfo{booktitle}{\emph{2020 IEEE 36th international conference on data engineering (ICDE)}}. IEEE, \bibinfo{pages}{841--852}.
\newblock


\bibitem[Liu et~al\mbox{.}(2021)]%
        {liu2021imbens}
\bibfield{author}{\bibinfo{person}{Zhining Liu}, \bibinfo{person}{Jian Kang}, \bibinfo{person}{Hanghang Tong}, {and} \bibinfo{person}{Yi Chang}.} \bibinfo{year}{2021}\natexlab{}.
\newblock \showarticletitle{IMBENS: Ensemble class-imbalanced learning in Python}.
\newblock \bibinfo{journal}{\emph{arXiv preprint arXiv:2111.12776}} (\bibinfo{year}{2021}).
\newblock


\bibitem[Liu et~al\mbox{.}(2020b)]%
        {liu2020mesa}
\bibfield{author}{\bibinfo{person}{Zhining Liu}, \bibinfo{person}{Pengfei Wei}, \bibinfo{person}{Jing Jiang}, \bibinfo{person}{Wei Cao}, \bibinfo{person}{Jiang Bian}, {and} \bibinfo{person}{Yi Chang}.} \bibinfo{year}{2020}\natexlab{b}.
\newblock \showarticletitle{MESA: boost ensemble imbalanced learning with meta-sampler}.
\newblock \bibinfo{journal}{\emph{Advances in neural information processing systems}}  \bibinfo{volume}{33} (\bibinfo{year}{2020}), \bibinfo{pages}{14463--14474}.
\newblock


\bibitem[Liu et~al\mbox{.}(2023)]%
        {liu2023topological}
\bibfield{author}{\bibinfo{person}{Zhining Liu}, \bibinfo{person}{Zhichen Zeng}, \bibinfo{person}{Ruizhong Qiu}, \bibinfo{person}{Hyunsik Yoo}, \bibinfo{person}{David Zhou}, \bibinfo{person}{Zhe Xu}, \bibinfo{person}{Yada Zhu}, \bibinfo{person}{Kommy Weldemariam}, \bibinfo{person}{Jingrui He}, {and} \bibinfo{person}{Hanghang Tong}.} \bibinfo{year}{2023}\natexlab{}.
\newblock \showarticletitle{Topological Augmentation for Class-Imbalanced Node Classification}.
\newblock \bibinfo{journal}{\emph{arXiv preprint arXiv:2308.14181}} (\bibinfo{year}{2023}).
\newblock


\bibitem[Mehrabi et~al\mbox{.}(2021)]%
        {mehrabi2021survey}
\bibfield{author}{\bibinfo{person}{Ninareh Mehrabi}, \bibinfo{person}{Fred Morstatter}, \bibinfo{person}{Nripsuta Saxena}, \bibinfo{person}{Kristina Lerman}, {and} \bibinfo{person}{Aram Galstyan}.} \bibinfo{year}{2021}\natexlab{}.
\newblock \showarticletitle{A survey on bias and fairness in machine learning}.
\newblock \bibinfo{journal}{\emph{ACM computing surveys (CSUR)}} \bibinfo{volume}{54}, \bibinfo{number}{6} (\bibinfo{year}{2021}), \bibinfo{pages}{1--35}.
\newblock


\bibitem[Mroueh et~al\mbox{.}(2021)]%
        {mroueh2021fairmixup}
\bibfield{author}{\bibinfo{person}{Youssef Mroueh} {et~al\mbox{.}}} \bibinfo{year}{2021}\natexlab{}.
\newblock \showarticletitle{Fair mixup: Fairness via interpolation}. In \bibinfo{booktitle}{\emph{International Conference on Learning Representations}}.
\newblock


\bibitem[Mukherjee et~al\mbox{.}(2020)]%
        {mukherjee2020learnsim}
\bibfield{author}{\bibinfo{person}{Debarghya Mukherjee}, \bibinfo{person}{Mikhail Yurochkin}, \bibinfo{person}{Moulinath Banerjee}, {and} \bibinfo{person}{Yuekai Sun}.} \bibinfo{year}{2020}\natexlab{}.
\newblock \showarticletitle{Two simple ways to learn individual fairness metrics from data}. In \bibinfo{booktitle}{\emph{International Conference on Machine Learning}}. PMLR, \bibinfo{pages}{7097--7107}.
\newblock


\bibitem[Pan et~al\mbox{.}(2004)]%
        {pan2004rwr}
\bibfield{author}{\bibinfo{person}{Jia-Yu Pan}, \bibinfo{person}{Hyung-Jeong Yang}, \bibinfo{person}{Christos Faloutsos}, {and} \bibinfo{person}{Pinar Duygulu}.} \bibinfo{year}{2004}\natexlab{}.
\newblock \showarticletitle{Automatic multimedia cross-modal correlation discovery}. In \bibinfo{booktitle}{\emph{Proceedings of the tenth ACM SIGKDD international conference on Knowledge discovery and data mining}}. \bibinfo{pages}{653--658}.
\newblock


\bibitem[Paszke et~al\mbox{.}(2019)]%
        {paszke2019pytorch}
\bibfield{author}{\bibinfo{person}{Adam Paszke}, \bibinfo{person}{Sam Gross}, \bibinfo{person}{Francisco Massa}, \bibinfo{person}{Adam Lerer}, \bibinfo{person}{James Bradbury}, \bibinfo{person}{Gregory Chanan}, \bibinfo{person}{Trevor Killeen}, \bibinfo{person}{Zeming Lin}, \bibinfo{person}{Natalia Gimelshein}, \bibinfo{person}{Luca Antiga}, {et~al\mbox{.}}} \bibinfo{year}{2019}\natexlab{}.
\newblock \showarticletitle{Pytorch: An imperative style, high-performance deep learning library}.
\newblock \bibinfo{journal}{\emph{Advances in neural information processing systems}}  \bibinfo{volume}{32} (\bibinfo{year}{2019}).
\newblock


\bibitem[Pedregosa et~al\mbox{.}(2011)]%
        {pedregosa2011scikit}
\bibfield{author}{\bibinfo{person}{Fabian Pedregosa}, \bibinfo{person}{Ga{\"e}l Varoquaux}, \bibinfo{person}{Alexandre Gramfort}, \bibinfo{person}{Vincent Michel}, \bibinfo{person}{Bertrand Thirion}, \bibinfo{person}{Olivier Grisel}, \bibinfo{person}{Mathieu Blondel}, \bibinfo{person}{Peter Prettenhofer}, \bibinfo{person}{Ron Weiss}, \bibinfo{person}{Vincent Dubourg}, {et~al\mbox{.}}} \bibinfo{year}{2011}\natexlab{}.
\newblock \showarticletitle{Scikit-learn: Machine learning in Python}.
\newblock \bibinfo{journal}{\emph{the Journal of machine Learning research}}  \bibinfo{volume}{12} (\bibinfo{year}{2011}), \bibinfo{pages}{2825--2830}.
\newblock


\bibitem[Rosenblatt and Witter(2023)]%
        {rosenblatt2023cfasdp}
\bibfield{author}{\bibinfo{person}{Lucas Rosenblatt} {and} \bibinfo{person}{R~Teal Witter}.} \bibinfo{year}{2023}\natexlab{}.
\newblock \showarticletitle{Counterfactual fairness is basically demographic parity}. In \bibinfo{booktitle}{\emph{Proceedings of the AAAI Conference on Artificial Intelligence}}, Vol.~\bibinfo{volume}{37}. \bibinfo{pages}{14461--14469}.
\newblock


\bibitem[Ruggieri et~al\mbox{.}(2010a)]%
        {ruggieri2010discover}
\bibfield{author}{\bibinfo{person}{Salvatore Ruggieri}, \bibinfo{person}{Dino Pedreschi}, {and} \bibinfo{person}{Franco Turini}.} \bibinfo{year}{2010}\natexlab{a}.
\newblock \showarticletitle{Data mining for discrimination discovery}.
\newblock \bibinfo{journal}{\emph{ACM Transactions on Knowledge Discovery from Data (TKDD)}} \bibinfo{volume}{4}, \bibinfo{number}{2} (\bibinfo{year}{2010}), \bibinfo{pages}{1--40}.
\newblock


\bibitem[Ruggieri et~al\mbox{.}(2010b)]%
        {ruggieri2010dcube}
\bibfield{author}{\bibinfo{person}{Salvatore Ruggieri}, \bibinfo{person}{Dino Pedreschi}, {and} \bibinfo{person}{Franco Turini}.} \bibinfo{year}{2010}\natexlab{b}.
\newblock \showarticletitle{DCUBE: Discrimination discovery in databases}. In \bibinfo{booktitle}{\emph{Proceedings of the 2010 ACM SIGMOD International Conference on Management of data}}. \bibinfo{pages}{1127--1130}.
\newblock


\bibitem[Sander(2009)]%
        {lsa}
\bibfield{author}{\bibinfo{person}{Richard Sander}.} \bibinfo{year}{2009}\natexlab{}.
\newblock \bibinfo{title}{{Law School Admissions Dataset}}.
\newblock \bibinfo{howpublished}{Project SEAPHE}.
\newblock
\newblock
\shownote{http://www.seaphe.org/databases.php}.


\bibitem[Selbst et~al\mbox{.}(2019)]%
        {selbst2019socialfair}
\bibfield{author}{\bibinfo{person}{Andrew~D Selbst}, \bibinfo{person}{Danah Boyd}, \bibinfo{person}{Sorelle~A Friedler}, \bibinfo{person}{Suresh Venkatasubramanian}, {and} \bibinfo{person}{Janet Vertesi}.} \bibinfo{year}{2019}\natexlab{}.
\newblock \showarticletitle{Fairness and abstraction in sociotechnical systems}. In \bibinfo{booktitle}{\emph{Proceedings of the conference on fairness, accountability, and transparency}}. \bibinfo{pages}{59--68}.
\newblock


\bibitem[Speicher et~al\mbox{.}(2018)]%
        {speicher2018ge}
\bibfield{author}{\bibinfo{person}{Till Speicher}, \bibinfo{person}{Hoda Heidari}, \bibinfo{person}{Nina Grgic-Hlaca}, \bibinfo{person}{Krishna~P Gummadi}, \bibinfo{person}{Adish Singla}, \bibinfo{person}{Adrian Weller}, {and} \bibinfo{person}{Muhammad~Bilal Zafar}.} \bibinfo{year}{2018}\natexlab{}.
\newblock \showarticletitle{A unified approach to quantifying algorithmic unfairness: Measuring individual \&group unfairness via inequality indices}. In \bibinfo{booktitle}{\emph{Proceedings of the 24th ACM SIGKDD international conference on knowledge discovery \& data mining}}. \bibinfo{pages}{2239--2248}.
\newblock


\bibitem[Suresh and Guttag(2019)]%
        {suresh2019framework}
\bibfield{author}{\bibinfo{person}{Harini Suresh} {and} \bibinfo{person}{John~V Guttag}.} \bibinfo{year}{2019}\natexlab{}.
\newblock \showarticletitle{A framework for understanding unintended consequences of machine learning}.
\newblock \bibinfo{journal}{\emph{arXiv preprint arXiv:1901.10002}} \bibinfo{volume}{2}, \bibinfo{number}{8} (\bibinfo{year}{2019}).
\newblock


\bibitem[Tong et~al\mbox{.}(2006)]%
        {tong2006rwr}
\bibfield{author}{\bibinfo{person}{Hanghang Tong}, \bibinfo{person}{Christos Faloutsos}, {and} \bibinfo{person}{Jia-Yu Pan}.} \bibinfo{year}{2006}\natexlab{}.
\newblock \showarticletitle{Fast random walk with restart and its applications}. In \bibinfo{booktitle}{\emph{Sixth international conference on data mining (ICDM'06)}}. IEEE, \bibinfo{pages}{613--622}.
\newblock


\bibitem[Wang et~al\mbox{.}(2020)]%
        {wang2020robustfair}
\bibfield{author}{\bibinfo{person}{Serena Wang}, \bibinfo{person}{Wenshuo Guo}, \bibinfo{person}{Harikrishna Narasimhan}, \bibinfo{person}{Andrew Cotter}, \bibinfo{person}{Maya Gupta}, {and} \bibinfo{person}{Michael Jordan}.} \bibinfo{year}{2020}\natexlab{}.
\newblock \showarticletitle{Robust optimization for fairness with noisy protected groups}.
\newblock \bibinfo{journal}{\emph{Advances in neural information processing systems}}  \bibinfo{volume}{33} (\bibinfo{year}{2020}), \bibinfo{pages}{5190--5203}.
\newblock


\bibitem[Xu et~al\mbox{.}(2021)]%
        {xu2021robustfair}
\bibfield{author}{\bibinfo{person}{Han Xu}, \bibinfo{person}{Xiaorui Liu}, \bibinfo{person}{Yaxin Li}, \bibinfo{person}{Anil Jain}, {and} \bibinfo{person}{Jiliang Tang}.} \bibinfo{year}{2021}\natexlab{}.
\newblock \showarticletitle{To be robust or to be fair: Towards fairness in adversarial training}. In \bibinfo{booktitle}{\emph{International conference on machine learning}}. PMLR, \bibinfo{pages}{11492--11501}.
\newblock


\bibitem[Yan et~al\mbox{.}(2021)]%
        {yan2021dynamic}
\bibfield{author}{\bibinfo{person}{Yuchen Yan}, \bibinfo{person}{Lihui Liu}, \bibinfo{person}{Yikun Ban}, \bibinfo{person}{Baoyu Jing}, {and} \bibinfo{person}{Hanghang Tong}.} \bibinfo{year}{2021}\natexlab{}.
\newblock \showarticletitle{Dynamic knowledge graph alignment}. In \bibinfo{booktitle}{\emph{Proceedings of the AAAI conference on artificial intelligence}}, Vol.~\bibinfo{volume}{35}. \bibinfo{pages}{4564--4572}.
\newblock


\bibitem[Yurochkin et~al\mbox{.}(2020)]%
        {yurochkin2020sensr}
\bibfield{author}{\bibinfo{person}{Mikhail Yurochkin}, \bibinfo{person}{Amanda Bower}, {and} \bibinfo{person}{Yuekai Sun}.} \bibinfo{year}{2020}\natexlab{}.
\newblock \showarticletitle{Training individually fair ML models with sensitive subspace robustness}. In \bibinfo{booktitle}{\emph{International Conference on Learning Representations}}.
\newblock


\bibitem[Yurochkin and Sun(2021)]%
        {yurochkin2021sensei}
\bibfield{author}{\bibinfo{person}{Mikhail Yurochkin} {and} \bibinfo{person}{Yuekai Sun}.} \bibinfo{year}{2021}\natexlab{}.
\newblock \showarticletitle{SenSeI: Sensitive Set Invariance for Enforcing Individual Fairness}. In \bibinfo{booktitle}{\emph{International Conference on Learning Representations}}.
\newblock


\bibitem[Zemel et~al\mbox{.}(2013)]%
        {zemel2013lfr}
\bibfield{author}{\bibinfo{person}{Rich Zemel}, \bibinfo{person}{Yu Wu}, \bibinfo{person}{Kevin Swersky}, \bibinfo{person}{Toni Pitassi}, {and} \bibinfo{person}{Cynthia Dwork}.} \bibinfo{year}{2013}\natexlab{}.
\newblock \showarticletitle{Learning fair representations}. In \bibinfo{booktitle}{\emph{International conference on machine learning}}. PMLR, \bibinfo{pages}{325--333}.
\newblock


\bibitem[Zeng et~al\mbox{.}(2024)]%
        {zeng2024hierarchical}
\bibfield{author}{\bibinfo{person}{Zhichen Zeng}, \bibinfo{person}{Boxin Du}, \bibinfo{person}{Si Zhang}, \bibinfo{person}{Yinglong Xia}, \bibinfo{person}{Zhining Liu}, {and} \bibinfo{person}{Hanghang Tong}.} \bibinfo{year}{2024}\natexlab{}.
\newblock \showarticletitle{Hierarchical multi-marginal optimal transport for network alignment}. In \bibinfo{booktitle}{\emph{Proceedings of the AAAI Conference on Artificial Intelligence}}, Vol.~\bibinfo{volume}{38}. \bibinfo{pages}{16660--16668}.
\newblock


\bibitem[Zhang et~al\mbox{.}(2018)]%
        {zhang2018advfair}
\bibfield{author}{\bibinfo{person}{Brian~Hu Zhang}, \bibinfo{person}{Blake Lemoine}, {and} \bibinfo{person}{Margaret Mitchell}.} \bibinfo{year}{2018}\natexlab{}.
\newblock \showarticletitle{Mitigating unwanted biases with adversarial learning}. In \bibinfo{booktitle}{\emph{Proceedings of the 2018 AAAI/ACM Conference on AI, Ethics, and Society}}. \bibinfo{pages}{335--340}.
\newblock


\bibitem[Zhang and Wu(2017)]%
        {zhang2017cfanti}
\bibfield{author}{\bibinfo{person}{Lu Zhang} {and} \bibinfo{person}{Xintao Wu}.} \bibinfo{year}{2017}\natexlab{}.
\newblock \showarticletitle{Anti-discrimination learning: a causal modeling-based framework}.
\newblock \bibinfo{journal}{\emph{International Journal of Data Science and Analytics}}  \bibinfo{volume}{4} (\bibinfo{year}{2017}), \bibinfo{pages}{1--16}.
\newblock


\bibitem[Zhang et~al\mbox{.}(2017)]%
        {zhang2017cfremove}
\bibfield{author}{\bibinfo{person}{Lu Zhang}, \bibinfo{person}{Yongkai Wu}, {and} \bibinfo{person}{Xintao Wu}.} \bibinfo{year}{2017}\natexlab{}.
\newblock \showarticletitle{A Causal Framework for Discovering and Removing Direct and Indirect Discrimination}. In \bibinfo{booktitle}{\emph{Proceedings of the Twenty-Sixth International Joint Conference on Artificial Intelligence}}.
\newblock


\end{thebibliography}

\appendix

\section{Reproducibility}\label{sec:ap-rep}

\subsection{Dataset Statistics and Details}\label{sec:ap-rep-data}

\begin{table*}[t]
\caption{Dataset statistics. We report the quantity of samples with positive/negative labels, number of continuous features and discrete features (one-hot encoded), sensitive attribute used, as well as size and positive ratio of the privileged/protected group.}
\label{tab:dataset}
\vspace{-10pt}
\resizebox{0.9\textwidth}{!}{%
\begin{tabular}{c|c|c|c|c|c|c|c}
\hline
\multirow{2}{*}{\textbf{Dataset}} & \multirow{2}{*}{\textbf{Domain}} & \textbf{\#Samples} & \textbf{Positive} & \textbf{\#Features} & \textbf{Sensitive} & \textbf{Group Size} & \textbf{Positive Ratio} \\
 &  & \textbf{(Positive/Negative)} & \textbf{Ratio} & \textbf{(Cont./One-hot)} & \textbf{Attribute} & \textbf{(Privileged/Protected)} & \textbf{(Privileged/Protected)} \\ \hline
\multirow{2}{*}{\adult} & \multirow{2}{*}{census} & \multirow{2}{*}{11,208 / 34,014} & \multirow{2}{*}{24.8\%} & \multirow{2}{*}{5 / 93} & gender & 30,527 / 14,695 & 31.2\% / 11.4\% \\
 &  &  &  &  & race & 38,903 / 6,319 & 26.2\% / 15.8\% \\ \hline
\multirow{2}{*}{\compas} & \multirow{2}{*}{criminological} & \multirow{2}{*}{2,737 / 3,138} & \multirow{2}{*}{46.6\%} & \multirow{2}{*}{5 / 6} & sex & 4,714 / 1,161 & 49.2\% / 36.2\% \\
 &  &  &  &  & race & 3,528 / 2,347 & 51.3\% / 39.5\% \\ \hline
\lsa & educational & 15,482 / 39,966 & 27.9\% & 5 / 4 & race & 40,989 / 14,459 & 29.6\% / 23.1\% \\ \hline
\multirow{2}{*}{\meps} & \multirow{2}{*}{medical} & \multirow{2}{*}{2,721 / 13,118} & \multirow{2}{*}{17.2\%} & \multirow{2}{*}{5 / 120} & sex & 8,255 / 7,584 & 20.8\% / 13.3\% \\
 &  &  &  &  & race & 5,659 / 10,180 & 25.6\% / 12.5\% \\ \hline
\end{tabular}%
}
\end{table*}

\textbf{\adult dataset}
The \adult dataset~\citep{adult} contains census personal records with attributes like age, education, race, etc. 
The task is to determine whether a person makes over \$50K a year. 
\textbf{\compas dataset}
The \compas dataset~\citep{compas} is a criminological dataset recording prisoners' information like criminal history, jail and prison time, demographic, sex, etc. 
The task is to predict a recidivism risk score for defendants. 
\textbf{\lsa dataset}
\lsa (Law School Admission) dataset~\citep{lsa} contains admissions data from 25 law schools, features include applicant attributes like LSAT score, undergraduate GPA, residency, race, etc. 
The target label is the admission decision of each applicant.
\textbf{\meps dataset}
The \meps (Medical Expenditure Panel Survey) dataset~\citep{meps} comprises demographic features, health status, income, and other attributes of surveyed individuals. 
The goal is to predict whether the individual has high medical service utilization.
We use the AIF360~\citep{bellamy2019aif} toolbox\footnote{\url{https://github.com/Trusted-AI/AIF360}} to retrieve and process all used datasets.
Detailed data statistics are listed in Table~\ref{tab:dataset}.

\vspace{-6pt}
\subsection{Implementation Details}\label{sec:ap-rep-imp}

\paragraph{Baselines}
We implement the 10 FairML baselines using standard Python toolkits or official code base provided by the paper authors.
Specifically, we use the AIF360~\citep{bellamy2019aif} package\footnotemark[1] to implement \textbf{Reduction}, \textbf{FairReweight}, \textbf{LearnFairRep}, \textbf{AdvFair}; 
the fairlearn~\citep{bird2020fairlearn} package\footnote{\url{https://github.com/fairlearn/fairlearn}} for implementing \textbf{Threshold}; 
the inFairness~\citep{infairness} package\footnote{\url{https://github.com/IBM/inFairness}} for implementing \textbf{SenSR} and \textbf{SenSeI}; 
and the FBB benchmark\footnote{\url{https://github.com/ahxt/fair_fairness_benchmark}} for implementing \textbf{FairMixup}, \textbf{AdvFair}, and \textbf{HSIC}.
For \textbf{AdaFair}, we use the official code base\footnote{\url{https://github.com/iosifidisvasileios/AdaFair}} for implementation.
Their hyperparameters are fine-tuned to get the best fairness-utility trade-off.

\paragraph{Models}
We consider logistic regression and neural network as base models in our experiments.
We use scikit-learn~\citep{pedregosa2011scikit} to implement logistic regression.
DRO-based FairML methods (SenSR, SenSEI) that do not compatible with this pipeline are validated with neural networks implemented with PyTorch~\citep{paszke2019pytorch}.
For logistic regression, we use the default parameters specified by sklearn.
For neural network, we use the implementation provided in the official example in inFairness~\citep{infairness} for SenSR/SenSEI to guarantee fair comparison.
The neural network is a three-layer MLP with 100 hidden units and ReLU activation function.
We use Adam optimizer with learning rate 1e-3 and cross-entropy loss, and train the MLP for 100 epochs with a mini-batch size 32 until it is converged.

\paragraph{Hyperparameters}
\name bias attribution criterion itself does not have any hyperparameters.
Similarity computing involves three main parameters: numerical and categorical disparity thresholds $t_r$ and $t_d$ for determining sample comparability, and the damping factor $p$ of RWR used for computing similarity.
We choose $t_r=0.1$ and $t_d=2$ to offer sufficient comparable samples while maintaining sample semantic comparability.
The damping factor $p$ is set to $0.1$ (i.e., restart probability for RWR = 0.9) to guarantee locality while capturing global similarity.
In general, these three parameters can all be regarded as parameters that directly or indirectly determine the locality of similarity, the smaller they are, the more local the similarity is.
Specifically, smaller $t_r$/$t_d$ results in more strict comparability constraints (also fewer comparable sample pairs) and thus more locality, and smaller $p$ means higher restart probability in RWR, thus also more locality in similarity.
Overall, these three parameters do not affect the validity of similarity and AIM bias attribution.
Users can adjust these values (or use other expert-specified similarity metrics if available) based on the application scenario.

\begin{figure*}[t]
    \centering
    \includegraphics[width=\linewidth]{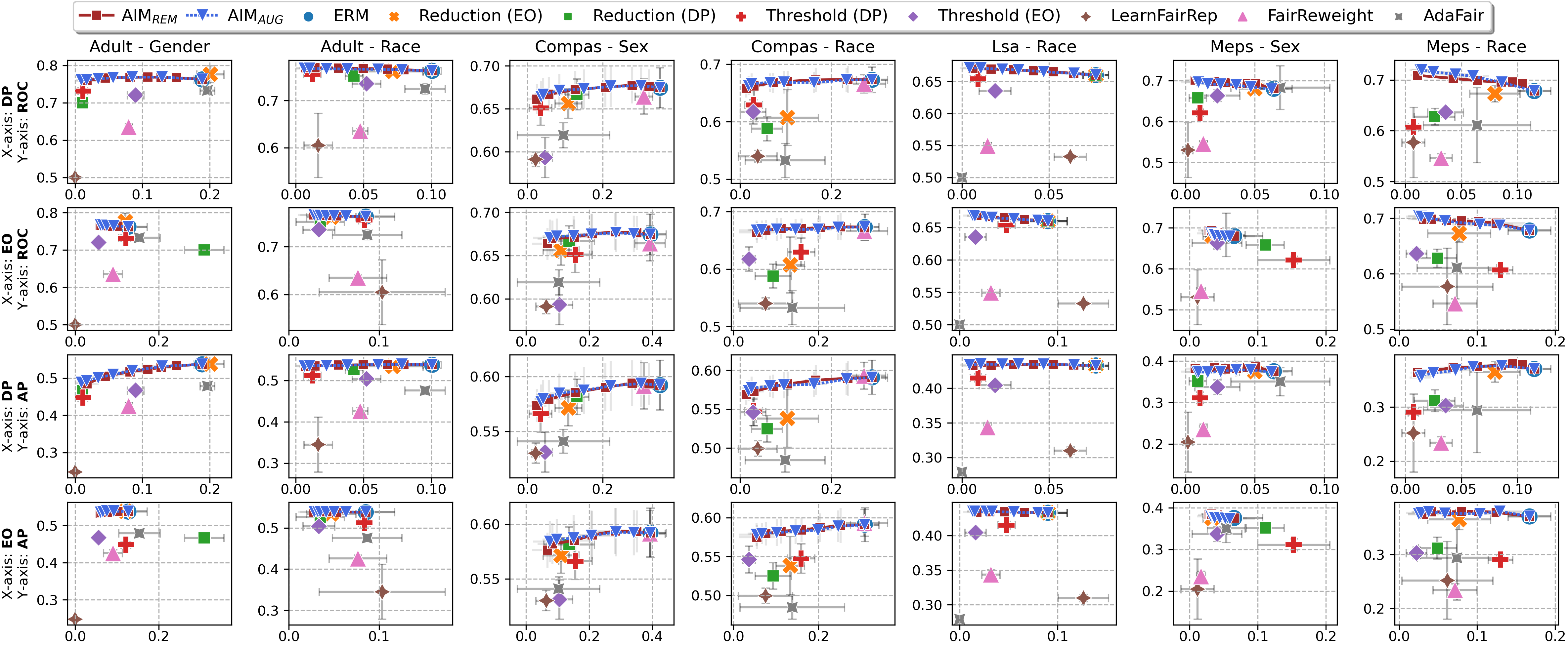}
    \vspace*{-20pt}
    \caption{
        Compare \namer and \namea with group fairness baselines.
        We show the utility-fairness trade-off between 2 utility metrics (x-axis) and 2 unfairness metrics (y-axis) on 7 real-world FairML tasks.
        Results close to the \textit{upper-left corner have better trade-offs, i.e., with low unfairness (x-axis) and high utility (y-axis).}
        Each column corresponds to a FairML task, and each row corresponds to a utility-unfairness metric pair.
        As \name's utility-unfairness trade-off can be controlled by the sample removal/augmentation budget, we show its performance with line plots. 
        We show error bars for both utility and unfairness.
    }
    \vspace*{-10pt}
    \label{fig:comp-group-full}
\end{figure*}

\section{More Results and Discussions}\label{sec:ap-discussion}

\paragraph{Additional Results}
In Figure~\ref{fig:comp-group}, we report the results with only ROC on 4 FairML tasks due to space limitation.
Here we provide additional results showing the utility-fairness trade-off between 2 utility metrics (ROC, AP) and 2 unfairness metrics (DP, EO) on 7 real-world FairML tasks from different domains.
Other protocols are identical to Figure~\ref{fig:comp-group}.
We report the additional results in Figure~\ref{fig:comp-group-full}.
It can be observed that the additional results are consistent with the conclusions derived in the paper from Figure~\ref{fig:comp-group}.
Across all 28 (2 utility metrics x 2 unfairness metrics x 7 FairML tasks) settings, \name achieves the optimal trade-off compared to other group fairness baselines: it either outperforms or matches the best baseline in terms of utility-fairness trade-off (close to the upper-left corner).
This further validates the effectiveness and generality of AIM in mitigating unfairness.

\paragraph{Complexity Analysis}
The complexity of estimating bias/credibility through \eqref{eq:estbias}/\eqref{eq:estcred} is $O(N^2)$. 
However, we note that both bias and credibility computations can be performed in matrix form, and thus can be efficiently accelerated with the parallel computing capabilities of modern GPUs.
The time complexity of bias computation can be reduced to $O(\frac{N^2}{C})$, where $C$ is the number of available computing units.
The complexity of computing similarity mainly arises from the matrix inversion step in RWR. 
There are many techniques can be employed to accelerate the solution of RWR, such as Fast Random Walk with Restart~\citep{tong2006rwr} that use low-rank approximation and Sherman–Morrison Lemma to approximate $(1-p\Tilde{\mW})^{-1}$.
In practice, if the dataset is large enough, and there are sufficient comparable samples to support the bias attribution and explanation for each sample, one can also consider directly using sample comparability as similarity to avoid additional computational costs.

\paragraph{Limitation and Future Works}
At the end of the paper, we discuss the limitations of AIM and possible future directions to address them.
One potential limitation is the static data assumption of AIM: as society evolves and relevant laws improve, the distribution of observed data also changes~\citep{DBLP:conf/www/Fu0MCBH23,DBLP:conf/kdd/FuFMTH22,DBLP:conf/kdd/FuZH20}.
It may not be reasonable to assess and interpret bias in a new sample using data collected a decade ago.
A possible solution is to introduce a reasonable time-discounting factor into the definitions of bias and credibility to account for concept drift~\citep{DBLP:conf/sigir/FuH21}. 
Additionally, in streaming data scenarios, the current definition requires recalculating similarity and sample bias/credibility when new data arrives.
Possible directions include maintaining a core matrix based on matrix low-rankness~\citep{DBLP:conf/cikm/FuXLTH20} to estimate data similarity after incorporating new data, thereby avoiding the need for re-computing RWR similarity each time. 
Exploring how to extend AIM to online scenarios to detect bias in newly arrived data in real-time, and its impact on existing data, would be a valuable future direction.
We also note that our class-imbalance-aware design significantly contributes to AIM's advantage in predictive utility. 
And the class imbalance problem (classifier's uneven attention to different classes)~\citep{liu2020self,liu2020mesa,liu2021imbens} is closely related to unfairness, particularly group fairness. Exploring the relationship between these issues and developing joint solutions will be a promising direction for future research.
Finally, we note that the comparable graph in this work can be seen as a simple nearest-neighbor-based graph. 
In many practical applications, there are complex networks of relationships between data points, forming graphs with intricate topologies~\citep{liu2023topological,zeng2024hierarchical,yan2021dynamic}.
FairML on graph data has recently gained significant attention, and extending \name form artificial nearest-neighbor-based graph to natural complex graph data would also be an interesting direction.

\end{document}